\documentclass{article}

\newtheorem{theorem}{Theorem}

\newtheorem{remark}{Remark}
\newtheorem{lemma}{Lemma}

\newtheorem{prop}{Proposition}
\newtheorem{assumption}{Assumption}

\newcommand{\gr}{\nabla}

\newcommand{\sgr}{\widetilde{\nabla}}

\newcommand{\w}{\bm{w}}

\newcommand{\E}{\mathbb{E}}

\usepackage[T1]{fontenc}
\usepackage[cmex10]{amsmath}

\usepackage{pgfplots}
\makeatletter
\def\BState{\State\hskip-\ALG@thistlm}
\makeatother

\usepackage[T1]{fontenc}
\usepackage[utf8]{inputenc}
\usepackage{pgfplots}
\usepackage{graphicx}
\usepackage{subcaption}
\usepackage{pstricks}
\usepackage{color}
\usepackage{amsmath}
\usepackage{bbm}
\usepackage{tcolorbox}
\usepackage{tikz}
\usepackage{pgfplots}
\usepackage{wrapfig}
\usepackage{lipsum}  
\usetikzlibrary{positioning}
\usepackage{tcolorbox}
\usepackage{amsfonts,amssymb}

\usepackage{natbib}
\usepackage{mathtools}
\usepackage{commath}
\usepackage{relsize}
\usepackage{bbm}
\usepackage{bm}
\usepackage[font={small}]{caption}
\usepackage{comment,color,soul}
\usepackage{enumerate} 
\usetikzlibrary{arrows,automata}
\usepackage{amsfonts}
\usepackage{url}
\usepackage{lipsum}
\usepackage[thinlines]{easytable}
\usepackage{nicefrac}
\usepackage{mysymbol}
\usepackage{float}
\usepackage{titling}
\usepackage{comment}
\usepackage[top=2.5cm,bottom=3.5cm,left=2.5cm,right=2.5cm,marginparwidth=1.75cm]{geometry}

 \makeatletter
\def\BState{\State\hskip-\ALG@thistlm}
\makeatother

\makeatletter
\newcommand{\algmargin}{\the\ALG@thistlm}
\makeatother

\makeatletter
\def\Let@{\def\\{\notag\math@cr}}
\makeatother

\usepackage[hidelinks]{hyperref}
\definecolor{darkred}{RGB}{150,0,0}
\definecolor{darkgreen}{RGB}{0,150,0}
\definecolor{darkblue}{RGB}{0,0,150}
\hypersetup{colorlinks=true, linkcolor=darkred, citecolor=darkblue, urlcolor=darkblue}

\usepackage[ruled]{algorithm2e}
\usepackage{enumitem}
\title{\textbf{Straggler-Resilient Federated Learning:\\ Leveraging the 
Interplay Between Statistical Accuracy and System Heterogeneity} \vspace{4mm}}

\date{}

\makeatletter
\renewcommand*{\@fnsymbol}[1]{\ensuremath{\ifcase#1\or 1 \or 2 \or 3 \or 4 \else\@ctrerr\fi}}
\makeatother
    
\newcommand*\samethanks[1][\value{footnote}]{\footnotemark[#1]}

\author{
Amirhossein Reisizadeh\thanks{UC Santa Barbara, \{\texttt{reisizadeh@ucsb.edu, ramtin@ece.ucsb.edu}\}.} , Isidoros Tziotis\thanks{The University of Texas at Austin, \{\texttt{isidoros\_13@hotmail.com, mokhtari@austin.utexas.edu}\}.} , Hamed Hassani\thanks{University of Pennsylvania, \{\texttt{hassani@seas.upenn.edu}\}.} ,  \vspace{5mm}\\
Aryan Mokhtari\samethanks[2] , Ramtin Pedarsani\samethanks[1]
}

 \thanksmarkseries{arabic}
\begin{document}

%

%





\maketitle

\begin{abstract}
\emph{Federated Learning} is a novel paradigm that involves learning from data samples distributed across a large network of clients while the data remains local. It is, however, known that federated learning is prone to multiple system challenges including system heterogeneity where clients have different computation and communication capabilities. Such heterogeneity in clients' computation speeds has a negative effect on the  scalability of federated learning algorithms and causes significant slow-down in their runtime due to the existence of stragglers. In this paper, we propose a novel straggler-resilient federated learning method that incorporates statistical characteristics of the clients' data to adaptively select the clients in order to speed up the learning procedure. The key idea of our algorithm is to start the training procedure with faster nodes and gradually involve the slower nodes in the model training once the statistical accuracy of the data corresponding to the current participating nodes is reached. The proposed approach reduces the overall runtime required to achieve the statistical accuracy of data of all nodes, as the solution for each stage is close to the solution of the subsequent stage with more samples and can be used as a warm-start. Our theoretical results characterize the speedup gain in comparison to standard federated benchmarks for strongly convex objectives, and our numerical experiments also demonstrate significant speedups in wall-clock time of our straggler-resilient method compared to federated learning benchmarks.

\end{abstract}

\section{Introduction}
 
Federated learning is a distributed learning framework whose objective is to train a learning model using the data of many clients (nodes) while keeping each node's data localized. In contrast with central learning or learning at a data center, the federated learning architecture  allows for preserving the clients' privacy as well as reducing the communication burden caused by transmitting data to a cloud. Nevertheless, as we move towards deploying federated learning in practice, it is becoming apparent that several major challenges still remain and the existing frameworks need to be rethought to address them. Important among these challenges is system (device) heterogeneity due to existence of \emph{straggling nodes} -- slow nodes with low computational capability -- that significantly slow down the model training \citep{li2019federated,kairouz2019advances}.

In this paper, we focus on system heterogeneity in federated learning frameworks and we leverage the interplay between statistical accuracy and system heterogeneity to design a straggler-resilient federated learning method
that carefully and adaptively selects a subset of available nodes in each round of training. A typical federated learning network consists of thousands of devices with a wide range of computational, communication, battery power, and storage characteristics. Hence, deploying traditional federated learning algorithms such as \texttt{FedAvg} \citep{mcmahan2017communication} on such a highly heterogeneous cluster of devices results in significant and unexpected delays due to existence of slow clients or stragglers. In most of such algorithms, \emph{all} the \emph{available} clients participate in the model training --regardless of their computational capabilities. Consequently, in each communication round of such methods, the server has to wait for the slowest node to complete its local updates and upload its local model which significantly slows down the training process. 

In this work, we aim to mitigate the effect of stragglers in federated learning frameworks based on an adaptive node participation approach in which clients are selected to participate in different stages of the training according to their computation speed.  We call our straggler-resilient proposal a \textbf{F}ederated \textbf{L}earning method with \textbf{A}daptive \textbf{N}ode \textbf{P}articipation algorithm or \texttt{FLANP}. The key idea of this scheme is to start the model training procedure with only a few clients which are the fastest among all the nodes. These participating clients continue to train their shared models while interacting with the parameter server. Note that since the server waits only for the participating nodes, it takes a short time for the participating (and fast) clients to promptly train a shared model. This model is, however, not accurate as it is trained over only a fraction of samples. We now increase the number of participating clients and include the next fastest subset of nonparticipating nodes in the training. Note that the model trained  from the previous stage can be a warm-start initialization for the current stage.

To discuss our main idea more precisely, consider a federated network of $N$ available nodes each storing $s$ data samples and suppose that we start the learning procedure with only $m$ clients. Once we solve the empirical risk minimization (ERM) problem corresponding to $m\times s$ samples of these nodes up to its statistical accuracy, we  geometrically increase the number of participating nodes to $n=\alpha m$ where $\alpha>1$, by adding the next $n-m$ fastest clients in the network. By doing so, the new ERM problem that we aim to solve contains the samples from the previous stage as well as the samples of the newly participating nodes. Also, the solution for the ERM problem at the previous stage (with $m$ clients) could be used as a warm-start for the ERM problem at the current stage with $n=\alpha  m$ nodes--This is due to the fact that all samples are drawn from a common distribution, and as a result, the optimal solution of the ERM problem with less samples is not far from the optimal solution of the ERM problems with more samples, as long as the larger set contains the smaller set.

In the proposed \texttt{FLANP} algorithm, as time progresses, we gradually increase the number of participating clients until we reach the full training set and all clients are involved. Note that in this procedure the slower clients are only used towards the end of the learning process, where the model is already close to the optimal model of the aggregate loss. 
 Another essential observation is that since the model trained in previous rounds already has a reasonable statistical accuracy and this model serves as the initial point of the next round of the iterative algorithm, the slower nodes are only needed to contribute in the final rounds of training, leading to a smaller wall-clock time.  This is in contrast with having all nodes participate in training from the beginning, which leads to computation time of each round being determined by the slowest node. In this paper, we formally characterize the gain obtained by using the proposed adaptive node participation scheme compared to the case that all available nodes contribute to training at each round. Next, we state a summary of our main contributions:



\vspace{-2mm}
\begin{itemize}
    \item We present a straggler-resilient federated learning meta-algorithm that leverages the interplay between statistical accuracy and device heterogeneity by adaptively activating heterogeneous clients.

    \item We specify the proposed meta-algorithm with a federated learning subroutine and present its optimization guarantees for strongly convex risks. Moreover, we characterize the wall-clock time of the proposed straggler-resilient scheme and demonstrate analytically that it achieves up to $\ccalO(\log(Ns))$ speedup gain compared to standard federated benchmarks.
    
    \item Our  numerical results show that our framework significantly improves the  wall-clock time compared to federated learning benchmarks --with either full or partial node participation-- for both convex and non-convex risks.
\end{itemize}

\paragraph{Related Work.}  System (device) heterogeneity challenge, which refers to the case that clients have different computational, communication and storage characteristics, has been studied in the literature. Asynchronous methods have demonstrated significant improvements in distributed data centers, however, such methods are less desirable in federated learning frameworks as they rely on bounded staleness of slow clients \citep{stich2019local,xie2019asynchronous}. The active sampling approach is another direction  in which the server aims for aggregating as many local updates as possible within a predefined time span \citep{nishio2019client}. More recently, \cite{wang2020tackling} proposed a normalized averaging method to mitigate stragglers in federated systems and the objective inconsistency due to mismatch in clients' local updates. Deadline-based computation has also been proposed to mitigate stragglers in decentralized training \citep{reisizadeh2019robust}. In a  different yet related direction, various federated learning algorithms have been studied to address the heterogeneity in clients' \emph{data} distributions \citep{karimireddy2019scaffold, haddadpour2020federated, li2018federated, reisizadeh2020robust,mohri2019agnostic,reddi2020adaptive}.

The idea of adaptive sample size training in which we solve a sequence of geometrically increasing ERM problems has been used previously for solving large-scale ERM problems. In particular, it has been shown that this scheme significantly improves the overall computational complexity of both first-order \citep{mokhtari2017first,mokhtari2019efficient} and second-order \citep{mokhtari2016adaptive,eisen2018large,jahani2020efficient} methods for achieving the statistical accuracy of the full training set. In this paper, we exploit this idea to develop \texttt{FLANP} for a completely different setting where we aim to address the issue of device heterogeneity in federated learning. 

As mentioned earlier, \texttt{FLANP} is a general meta-algorithm that can be employed with any choice of federated learning subroutine such as the celebrated \texttt{FedAvg} \citep{mcmahan2017communication} and its variants \citep{wang2018cooperative,li2019convergence,huo2020faster,pathak2020fedsplit,malinovsky2020local,wang2019adaptive}. Convergence properties of such methods have also been extensively studied in the literature \citep{zhou2017convergence,haddadpour2019convergence,haddadpour2019local,bayoumi2020tighter, stich2019error,wang2018adaptive,koloskova2020unified, reisizadeh2020fedpaq,liang2019variance}. In this paper, we only showcase the gain obtained by combining \texttt{FLANP} and \texttt{FedGATE} proposed in \citep{haddadpour2020federated}.

\section{Federated Learning Setup} \label{sec:setting}

In this section, we state our setup.
Consider a federated architecture where  $N$ nodes interact with a central server, and each node $i \in [N] = \{1,\cdots,N\}$ has access to $s$ data samples denoted by $\{z^i_1, \cdots, z^i_s\}$. These samples are drawn at the beginning of the training process, and nodes cannot draw new samples during training. 
Further, define  $\ell(\cdot, \cdot): \mathbb{R}^d \times \mathcal{Z}\to \mathbb{R} $ as a loss function where $\ell(\w, z^i_j)$ indicates how well the model $\w$ performs with respect to the sample $z^i_j$. Also, define the empirical loss of node $i$ as
\begin{align}
    L^i (\w) \coloneqq \frac{1}{s} \sum_{j=1}^{s} \ell(\w, z^i_j). \nonumber
\end{align}
For any $1 \leq n \leq N$, we denote by $L_n (\w)$ the collective empirical risk corresponding to samples of all nodes $\{1,\cdots,n\}$, which is defined as
\begin{align} \label{eq:L_n}
    L_n (\w) \coloneqq \frac{1}{n} \sum_{i=1}^{n} L^i (\w).
\end{align}
In other words, $L_n(\w)$ represents the average loss over the $n \times s$ samples stored at nodes $\{1,\cdots,n\}$. We let $\w^*_n$ denote the optimal minimizer of the loss $L_n(\w)$, i.e., $\w^*_n = \argmin_{\w} L_n(\w)$.

We assume that the samples $z^i_j$ are i.i.d. realizations of a random variable $Z$ with probability distribution $\ccalP$. The problem of finding a global model for the aggregate loss of all available $N$ nodes, which can be considered as the empirical risk minimization (ERM) in \eqref{eq:L_n} for $n=N$, i.e., 
\begin{equation} \label{eq: erm}
   \!\! \min_{\w}  L_N(\w)\!=\! \frac{1}{N} \sum_{i=1}^{N} L^i (\w) \!=\! \frac{1}{Ns} \sum_{i=1}^{N} \sum_{j=1}^{s} \ell(\w, z^i_j) 
\end{equation}
is a surrogate for the expected risk minimization
\begin{align}
   \min_{\w} L (\w) \coloneqq \E_{Z \sim \ccalP} [\ell(\w, Z)]. \nonumber
\end{align}
Note that our ultimate goal is to find the optimal solution of the expected risk $\w^* = \argmin_{\w} L(\w)$; however, since the underlying distribution $\ccalP$ is unknown and we only have access to a finite number of realizations of the random variable $Z$, i.e., $\{z^i_1, \cdots, z^i_s\}_{i=1}^N$, we settle for minimizing the ERM problem in~\eqref{eq: erm}. 

We also note that the described setup is slightly different from the prior work on data homogeneous and data heterogeneous settings. To be more precise, in the standard data homogeneous settings the samples of all nodes are drawn according to a common distribution, which is similar to our setup. However, in homogeneous settings, it is also assumed that each node can draw new i.i.d. samples at each round, which is different from our setting where all samples are selected at the first round. Indeed, this difference is crucial, as in the described homogeneous settings the samples used at each round are independent and the local gradient direction of each node is an unbiased estimator of the aggregate loss, but in our setting we cannot guarantee that the samples used at each round are independent and more importantly we cannot ensure that the local gradient direction $\nabla L^i$ is an unbiased estimator for the aggregate loss gradient $\nabla L_N$. Considering this observation, our setup might be closer to the data heterogeneous setting, since local gradients are not aligned with each other (in expectation). However, our setup is less general compared to a \emph{data} heterogeneous setting, as we still assume that the given samples are all drawn from a common distribution, unlike the data heterogeneous setting where nodes could have arbitrary different data distributions.


\noindent \textbf{Statistical Accuracy.} The difference of expected and empirical risks $L_n(\w) - L(\w)$ is referred to as the estimation error and can be bounded by a function of the sample size. In particular, since $L_n(\w)$ captures $ns$ samples, we assume that there exists a constant $V_{ns}$ bounding the estimation error with high probability, 
\begin{align} 
    \sup_{\w} |L_n(\w) - L(\w)| \leq V_{ns}, \quad \text{w.h.p.} \nonumber
\end{align}
The estimation error $V_{ns}$ has been deeply studied in the statistical learning literature \citep{vapnik2013nature,bousquet2002concentration}. In particular, it has been shown that for strongly convex functions the estimation error is proportional to the inverse of sample size \citep{bartlett2006convexity,frostig2015competing}. In this work, we also assume that $V_{ns} = \frac{c}{ns}$ for a constant $c$. Note that for the loss function $L_n$ once we find a point $\tilde{\w}$ that has an optimization error of $V_{ns}$, i.e., $L_n(\tilde{\w})-L_n(\w_n^*)\leq V_{ns}$, there is no gain in improving the optimization error as the overall error with respect to the expected risk $L$ would not improve. Hence, when we find a point $\tilde{\w}$ such that $L_n(\tilde{\w})-L_n(\w_n^*)\leq V_{ns}$, we state that it has reached the statistical accuracy of $L_n$. Our goal is to find a solution $\w_N$ that is within the statistical accuracy of the ERM problem corresponding to the full training set defined in \eqref{eq: erm}.


\noindent \textbf{System Heterogeneity Model.}
As mentioned earlier, federated clients attribute a wide range of computational powers leading to significantly different processing time for a fixed computing task such as gradient computation and local model update. To be more specific, for each node $i \in [N]$, we let $T_i$ denote the (expected) time to compute one local model update. The time for such update is mostly determined by the computation time of a fixed batch-size stochastic gradient of the local empirical risk $L_i(\w)$. Clearly, larger $T_i$ corresponds to slower clients or stragglers. Without loss of generality, we assume that the nodes are sorted from faster to slower, that is, $T_1 \leq \cdots \leq T_N$ with node $1$ and $N$ respectively identifying the fastest and slowest nodes in the network.

\section{Adaptive Node Participation Approach} \label{sec: method}

Several federated learning algorithms have been proposed to solve the ERM problem in \eqref{eq: erm} such as \texttt{FedAvg} \citep{mcmahan2017communication}, \texttt{FedProx} \citep{li2018federated},
\texttt{SCAFFOLD} \citep{karimireddy2019scaffold}, \texttt{DIANA} \citep{mishchenko2019distributed},
\texttt{FedGATE} \citep{haddadpour2020federated}, etc. These methods consist of many rounds of local computations by the local clients and communication with the parameter server. Alas, in all such approaches, \emph{all the available  nodes} in the network --regardless of their computational capabilities-- contribute to model learning throughout the entire procedure. As explained in the previous section, federated clients operate in a wide range of computational characteristics, and therefore, the parameter server has to wait for the slowest node in each communication round to complete its local computation task. All in all, the slowest nodes (stragglers) determine the overall wall-clock time of the such federated algorithms which causes significant slow-down.

In this section, we first describe our proposed approach to mitigate stragglers in federated learning settings and lay out the rational behind our method. Our proposal, \texttt{FLANP}, is essentially a meta-algorithm that can be specified with the choice of any particular federated learning subroutine. The rest of the section focuses on a particular case of the proposed \texttt{FLANP} algorithm where the federated learning subroutine is picked to be \texttt{FedGATE} proposed by \cite{haddadpour2020federated}.

\subsection{\texttt{FLANP}: A Straggler-Resilient Federated Learning Meta-Algorithm}

Consider the federated learning setting described in Section \ref{sec:setting} consisting of $N$ available nodes with different computation times $T_1 \leq \cdots \leq T_N$. Our proposal to address the device heterogeneity and mitigate the stragglers is as follows. We start off the learning procedure with the $n_0$ fastest nodes, that are nodes $\{1, 2, \cdots, n_0\}$, and $n_0$ is much smaller than the total number of available nodes $N$. Using \texttt{Federated\_Solver} which is a federated learning subroutine of choice, e.g., \texttt{FedAvg} or \texttt{FedGATE}, the set of $n_0$ participating nodes proceed to minimize the empirical risk corresponding to their data points, which we denote by $L_{n_0}(\w)$ as defined in \eqref{eq:L_n}. This continues until the $n_0$ nodes reach their corresponding statistical accuracy, that is, they reach a global model $\w_{n_0}$ such that $L_{n_0}(\w_{n_0}) - L_{n_0}(\w_{n_0}^*) \leq V_{n_0 s}$. Note that at this stage the server has to wait only for the slowest client among the participating ones, i.e., node $n_0$, which is potentially much faster than the network's slowest node $N$.

\begin{algorithm}[t]
\textbf{Initialize} fast-to-slow nodes $\{1,\cdots,N\}$, $n \!=\! n_0$ participating nodes with initial global model $\w_{n_0}$

\While{$n \leq N$}{
    \While{$L_{n}(\w_{n}) - L_{n}(\w_{n}^*) > V_{n s}$}{
        nodes $\{1,\cdots,n\}$ are participating and update local models via \texttt{Federated\_Solver} \\
        server aggregates local models from nodes $\{1,\cdots,n\}$ and updates global model $\w_{n}$
    }
    $n \gets \min\{2n,  N\}$ \hfill{\color{darkblue}{\texttt{\% doubling  the participants}}}
}
\caption{\texttt{FLANP}}\label{alg: robust meta-alg}
\end{algorithm}

Per our discussion in Section \ref{sec:setting}, a more accurate solution than $\w_{n_0}$ would not help improving the optimality gap. Therefore, once statistical accuracy is achieved, the procedure is terminated and we increase the number of participating nodes from $n_0$ to $2 n_0$, implying that nodes $\{1, 2, \cdots, 2n_0\}$ now participate in improving the global model. Using the model learned form the previous stage $\w_{n_0}$ as the initial model, the nodes in 
$\{1, 2, \cdots, 2n_0\}$ employ the solver \texttt{Federated\_Solver} and continue the learning procedure to reach a global model $\w_{2n_0}$ within their corresponding statistical accuracy. That is, the global model $\w_{2n_0}$ satisfies $L_{2n_0}(\w_{2n_0}) - L_{2n_0}(\w_{2n_0}^*) \leq V_{2n_0 s}$. In this stage with $2n_0$ participating nodes, the computation delay is determined by the slowest participating node, i.e., node $2n_0$, which is slower than the previous stage with $n_0$ participating nodes, but faster than the network's slowest node, $N$. The  procedure of geometrically increasing the number of participating nodes continues till the set of participating nodes contains all the available $N$ nodes and nodes find the final global model $\w_{N}$ within the statistical accuracy of the global loss function $L_N(\w)$. Algorithm \ref{alg: robust meta-alg} summarises the straggler-resilient meta-algorithm described above.

From a high-level perspective, Algorithm \ref{alg: robust meta-alg} exploits the faster nodes in the beginning of the learning procedure to promptly reach a global model withing their statistical accuracy. By doing so, the parameter server avoids waiting for the slow nodes to complete their local updates; however, the optimality gap of such models are relatively large since only a fraction of data samples have contributed in the global model. By gradually increasing the number of participating nodes and activating slower nodes, the quality of the global model improves while the synchronous computation slows down due to slower nodes. The key point is that slower nodes join the learning process towards the end of process.

The criterion in Algorithm \ref{alg: robust meta-alg}, that is $L_{n}(\w_{n}) - L_{n}(\w_{n}^*) > V_{n s}$, verifies that the current global model satisfies the statistical accuracy corresponding to $n$ participating nodes $\{1,\cdots,n\}$. This condition, however, is not easy to check since nodes do not have access to the optimal solution $\w_n^*$. A sufficient and computationally feasible criterion is to check if $\Vert \gr L_n (\w_n) \Vert^2 \leq 2 \mu V_{n s}$, when the loss function $\ell$ is $\mu$-strongly convex.

Now consider a subset of $m$ participating nodes which we denote by $\ccalN_m \subseteq [N]$ and assume that they have reached their statistical accuracy, i.e., model $\w^*_m$ such that $L_{m}(\w_{m}) - L_{m}(\w_{m}^*) \leq V_{m s}$. Next proposition characterizes the initial suboptimality when the model  $\w^*_m$ is used as the initial warm-start model for a larger set of participating nodes $\ccalN_n$ where $\ccalN_m \subseteq \ccalN_n$.
\begin{prop} \label{prop}
Consider two subsets of nodes $\ccalN_m \subseteq \ccalN_n$ and assume that model $\w_m$ attains the statistical accuracy for the empirical risk associated with nodes in $\ccalN_m$, that is, $\Vert \gr L_m (\w_m) \Vert^2 \leq 2 \mu V_{m s}$ where the loss function $\ell$ is $\mu$-strongly convex. Then the suboptimality of $\w_m$ for risk $L_n$, i.e., $L_n(\w_m) - L_n(\w^*_n)$ is w.h.p. bounded above as follows
\begin{align} \label{eq: L_n - L^*_n}
    L_n(\w_m) - L_n(\w^*_n) 
    &\leq
    \frac{2(n - m)}{n} \left( V_{(n-m)s} + V_{ms} \right) + V_{ms}. \nonumber
\end{align}
\end{prop}

Proposition \ref{prop} demonstrates that a model attaining the statistical accuracy for $m$ nodes in $\ccalN_m$ can be used as an initial warmup model for the ERM corresponding to $n$ nodes in $\ccalN_n$ which includes $\ccalN_m$. In particular, when the number of participating nodes is doubled, i.e., $n = 2m$, then the initial sub-optimality error is bounded above by $L_n(\w_m) - L_n(\w^*_n) \leq 3 V_{ms}$.

As the proposed meta-algorithm \texttt{FLANP} in Algorithm~\ref{alg: robust meta-alg} provides a general mechanism to mitigate stragglers in federated settings, one needs to specify the inner optimization subroutine \texttt{Federated\_Solver} to quantify the speedup of the proposed approach. This subroutine may be any federated learning algorithm (with possible adjustments). Next, we pick the subroutine in Algorithm \ref{alg: robust meta-alg} to be \texttt{FedGATE} \citep{haddadpour2020federated}.

\subsection{\texttt{FLANP} via \texttt{FedGATE}}

\begin{algorithm}[t]
\textbf{Initialize} $n = n_0$ participating nodes, initial model $\w_{n_0}$, initial gradient tracking $\delta_i^{(0)} = 0$ for participating nodes $i \in \{1,\cdots,n_0\}$

\While{$n \leq N$}{

    $r=0$ \hfill{\color{darkblue}{\texttt{\% reset round counter for each stage}}}
    
    \For{\normalfont{participating nodes} $i \in \{1,\cdots,n\}$}{
    $\delta_i^{(0)} = 0$ \hfill{\color{darkblue}{\texttt{\% reset gradient tracking}}}
        }
    
    \While{$\Vert \gr L_n (\w_n) \Vert^2 > 2 \mu V_{n s}$}{
        \For{\normalfont{participating nodes} $i \in \{1,\cdots,n\}$}{
        
            $\w^{(0,r)}_i = \w_n$
            
            \For{$c = 0, \cdots, \tau_n-1$}{
                set $d^{(c,r)}_i = \sgr L^i (\w^{(c,r)}_i) - \delta^{(r)}_i$
                
                update $\w^{(c+1,r)}_i = \w^{(c,r)}_i - \eta_n d^{(c,r)}_i$
            }
            send $\Delta^{(r)}_i = (\w_n - \w^{(\tau_n,r)}_i) / \eta_n$ to server
            
            update $\delta^{(r+1)}_i = \delta^{(r)}_i + \frac{1}{\tau_n} (\Delta^{(r)}_i - \Delta^{(r)})$
        }
        server computes and broadcasts $\Delta^{(r)} = \frac{1}{n} \sum_{i=1}^{n} \Delta^{(r)}_i$
        
        server updates and broadcasts global model $\w_n \gets \w_n - \eta_n \gamma_n \Delta^{(r)}$
        
        participating nodes $i \in \{1,\cdots,n\}$ upload gradients $\gr L^i (\w_n)$ to  server
        
        $r \gets r + 1$
    }
    $n \gets \min\{2n,  N\}$ \hfill{\color{darkblue}{\texttt{\% doubling the participants}}}
}
\caption{\texttt{FLANP} via \texttt{FedGATE}}\label{alg: straggler-robust fed}
\end{algorithm}

We now focus on a specific instance of the meta-algorithm proposed in Algorithm \ref{alg: robust meta-alg} where the subroutine \texttt{Federated\_Solver} is set to be \texttt{FedGATE}, a federated learning algorithm that employs gradient tracking variables to provide tight convergence guarantees for nodes with heterogeneous data distributions. One reason that we use \texttt{FedGATE} as a subroutine is that it can handle the case that local gradients are not an unbiased estimator of the global loss gradient, which is the case in our setting. Algorithm \ref{alg: straggler-robust fed} demonstrates how the adaptive node participation approach in \texttt{FLANP} is adopted to mitigate straggler delays in \texttt{FedGATE}.

We begin the first stage of Algorithm~\ref{alg: straggler-robust fed} with activating the $n=n_0$ fastest nodes $\{1,\cdots,n_0\}$ and initialize them with an arbitrary global model $\w_{n_0}$. We also reset the gradient tracking variables $\delta^{(0)}_i$ to be zero for all participating nodes at the beginning of each stage. Variables $\delta_i$ aim to correct the directions of local updates at node $i$ by tracking the difference of local stochastic gradients $\sgr L^i$ and global gradients $\gr L_n$ such that directions $d_i$ closely follow the correct global gradient direction. After $\tau_n$ iterations of local updates at any participating node in round $r$, accumulations of local gradients $\Delta^{(r)}_i$ are uploaded to the server where it updates the average $\Delta^{(r)}$ and the global model $\w_n$ using two stepsizes $\eta_n,\gamma_n$. Note that the stepsizes $\eta_n,\gamma_n$ are fixed throughout each stage with $n$ participating nodes but vary for different stages as $n$ increases. After updating the global model $\w_n$ at the end of each round, participating nodes upload their local gradients $\gr L^i(\w_n)$ such that the server aggregates and computes the global gradient $\gr L_n(\w_n)$ and checks whether the condition $\Vert \gr L_n (\w_n) \Vert^2 \leq 2 \mu V_{n s}$ is satisfied. After $R_n$ rounds of communications, this condition is satisfied and the set of $n$ participating nodes reach the global model $\w_n$ within their statistical accuracy. Therefore, we augment the set of participating nodes (from faster to slower ones) from $\{1,\cdots,n\}$ to $\{1,\cdots,2n\}$ leading to a new stage. The above procedure continues until the set of participating nodes contains all $N$ available nodes.

\section{Theoretical Results}

In this section, we provide rigorous analysis for the proposed straggler-resilient \texttt{FLANP} scheme outlined in Algorithm \ref{alg: straggler-robust fed}, which employs \texttt{FedGATE} as its subroutine. We first characterize optimization guarantees of Algorithm \ref{alg: straggler-robust fed}. Using such results, we derive the expected runtime of our proposed algorithm and the speedup gain it provides compared to naive methods.

\subsection{Optimization Guarantees}

In this section, we aim to characterize the required communication and computation  for solving each subproblem. To be more precise, consider the case that we are given a model $\w_m$ which is within the statistical accuracy of $L_m$ corresponding to $m$ fastest nodes, and the goal is to find a new model $\w_n$ that is within the statistical accuracy of $L_n$ corresponding to $n$ fastest nodes, where $n=2m$. To analyze this procedure, we need to quantify three parameters: the choice of stepsizes $\eta_n, \gamma_n$, the number of local updates $\tau_n$ at each participating node, and the number of communication rounds with the server $R_n$. Note that for all of these parameters we use the index $n$ as they refer to the case that $n$ nodes participate in learning. Before stating our result, we first formally state our assumptions.


\begin{assumption}\label{assumption loss}
The loss function $\ell(\w, z)$ is $\mu$-strongly convex with respect to $\w$. Moreover, the gradient $\gr_{\w} \ell (\w, z)$ is Lipschitz continuous with constant~$L$ and the condition number is $\kappa \coloneqq L/\mu$.
\end{assumption}
The conditions in Assumption \ref{assumption loss} imply that the empirical risks $L_n (\w)$ and local loss functions $L^i(\w)$ are $\mu$-strongly convex and have $M$-Lipschitz gradients.

As we discussed in Section \ref{sec:setting}, the approximation error between the expected and the empirical risks corresponding to $ns$ data samples can be bounded as $|L_n(\w) - L(\w)| \leq V_{ns}$ with high probability. Next, we formalize this assumption.
\begin{assumption}\label{assumption: approx error}
We assume that the approximation error for the expected loss $L(\w)$ using $ns$ samples of $n$ nodes in the empirical risk $L_n(\w)$ is upper-bounded as
\begin{equation} 
    \sup_{\w}\,\,  |L_n(\w) - L(\w)| \leq V_{ns}, \quad \text{w.h.p.} \nonumber
\end{equation}
where $V_{ns}=\mathcal{O}(1/ns)$. Moreover, we assume that the approximation error for gradients is upper-bounded by
\begin{align} 
    \sup_{\w} \,\, \norm{ \gr L_n(\w) - \gr L(\w)} \leq \sqrt{V_{ns}}, \quad \text{w.h.p.} \nonumber
\end{align}
\end{assumption}

We now turn our focus to the proposed Algorithm \ref{alg: straggler-robust fed}. 

\begin{theorem} \label{theorem:1}
Consider the federated ERM problem in~\eqref{eq: erm} and suppose Assumptions \ref{assumption loss} and \ref{assumption: approx error} hold. Let the proposed {\normalfont\texttt{FLANP}} in Algorithm \ref{alg: straggler-robust fed} be initialized  with the fastest $n_0$ nodes in $\{1,\cdots,n_0\}$ and the model $\w_{n_0}$. Moreover, suppose the variance of stochastic local gradients is bounded above by $\sigma^2$, i.e.,  $\E \Vert \sgr L^i(\w) - \gr L^i(\w) \Vert^2  \leq \sigma^2$ for all nodes $i$.
At any stage of Algorithm \ref{alg: straggler-robust fed} with $n$ participating nodes, if the stepsizes are chosen as 
$
    \eta_n = \frac{\alpha_n}{\tau_n \sqrt{n}},
    \gamma_n = \frac{\sqrt{n}}{2 \alpha_n L},
$
where
\begin{align}    
    \alpha_n 
    \leq
    \min \left\{ 
    \frac{1}{12\sqrt{3} \kappa \sqrt{\kappa L}},
    \frac{\sqrt{n}}{12 \sqrt{2(3\kappa^2 + 2) (2\mu + 1) \kappa L}},
    \left( \frac{ \sqrt{n}}{96 \kappa^2 L^2} \right)^{1/3}, 
    \frac{\sqrt{n}}{\sqrt{15 c \kappa L}} ,
    \frac{\sqrt{n}}{L \sqrt{30}} \right\}, \nonumber
\end{align}
and each node runs $\tau_n = 1.5 s \sigma^2 / c$ local updates, where  $c$ captures the constant term in the statistical accuracy $V_{ns} = \frac{c}{ns}$, then nodes reach the statistical accuracy of $L_n$ after $R_n = 12 \kappa \ln(6)$ rounds of communication. 
\end{theorem}

The result in Theorem \ref{theorem:1} guarantees that if we initialize Algorithm \ref{alg: straggler-robust fed} with $n_0$ fastest nodes and in each stage the participating nodes update their local models according to Algorithm \ref{alg: straggler-robust fed} for $\tau = \ccalO(s)$ iterations and $R = \ccalO(\kappa)$ rounds, before doubling the number of participating nodes, then at the end of the final stage in which all $N$ nodes are participating, we reach a model $\w_N$ that attains the statistical accuracy of the empirical risk $L_N(\w)$. More precisely, we have $\E [L_N(\w_N) - L_N(\w^*_N)] \leq V_{Ns}$. We would like to mention that to obtain the best guarantee, $\tau_n$ and $R_n$ are independent of number of participating nodes $n$, while the stepsizes $\eta_n$ and $\gamma_n$ change as the number of participating nodes increases.

\subsection{Wall-Clock Time Analysis}

We have thus far established convergence properties of Algorithm \ref{alg: straggler-robust fed} and precisely derived its parameters.  It is, however, equally important to show that it provably mitigates stragglers in a federated learning framework and hence speeds up the overall wall-clock time. In the following, we first characterize the running time of Algorithm \ref{alg: straggler-robust fed} and then compare it with the one for straggler-prone \texttt{FedGATE} benchmarks.

\begin{prop} \label{prop2}
The average runtime of Algorithm \ref{alg: straggler-robust fed} to reach the overall statistical accuracy of the ERM problem corresponding to all nodes (defined in  \eqref{eq: erm}) is 
\begin{align} \label{eq: T_FLANP}
    \E[T_{\normalfont{\texttt{FLANP}}}] = R \, \tau  \sum_{i \,=\, n_0, \, 2n_0, \, 4n_0,\cdots, \, N} T_i,
\end{align}
where $R = 12 \kappa \ln(6)$ and $\tau = 1.5 s \sigma^2 / c$ per Theorem \ref{theorem:1}.
\end{prop}

\emph{Proof.} \normalfont{As described in Section \ref{sec: method} and Algorithm \ref{alg: straggler-robust fed}, we start with the $n_0$ fastest nodes and according to Theorem \ref{theorem:1}, we run $R = 12 \kappa \ln(6)$ rounds of local updates with $\tau = 1.5 s \sigma^2 / c$ iterations per round. Since the iteration time in this stage is determined by the slowest node, i.e., node $n_0$, then the overall wall-clock time for this stage with $n_0$ participating nodes is $R \tau T_{n_0}$. The same argument holds for subsequent stages with $2n_0, 4n_0, \cdots, N$ participating nodes and the total learning time is derived as in \eqref{eq: T_FLANP}.}

The result in Proposition \ref{prop2} further demonstrates how the adaptive node participation approach incorporates faster nodes in order to save in the overall wall-clock time. As Theorem \ref{theorem:1} shows, it suffices for each participating node in the straggler-resilient Algorithm \ref{alg: straggler-robust fed} to run $R = \ccalO(\kappa)$ rounds of local updates and $\tau = \ccalO(s)$ iteration per round to reach the final statistical accuracy. Therefore, the overall wall-clock time of Algorithm~\ref{alg: straggler-robust fed} is order-wise $\ccalO(\kappa s (T_{n_0} + T_{2n_0} + \cdots + T_N))$.

To quantify the speedup gain provided by the proposed straggler-resilient method, we need to characterize the wall-clock time for the non-adaptive benchmark \texttt{FedGATE}. Note that in this benchmark, all the $N$ available nodes are participating in the training process from the beginning.

\begin{prop} \label{prop3}
The expected runtime for the non-adaptive benchmark {\normalfont \texttt{FedGATE}} to solve the federated ERM problem \eqref{eq: erm} and to reach the statistical accuracy of all the samples of the $N$ nodes is as follows
\begin{align} 
    \E[T_{\normalfont{\texttt{FedGATE}}}] = \ccalO(\kappa s \log(Ns) T_N), \nonumber
\end{align}
where $T_N$ denotes the unit computation time corresponding to the slowest node.
\end{prop}

As explained before, the total runtime for each round of local updates in a federated learning algorithm is determined by the slowest node as the parameter server has to wait for all nodes to finish their local updates. This is consistent with the result in Proposition \ref{prop3} as all $N$ nodes participate in training since the beginning of the algorithm, and hence the overall wall-clock time depends only on the slowest node with computation time $T_N$ which is the largest among $\{T_1,\cdots,T_N\}$. Note  that it can be easily verified that the overall runtime of \texttt{FLANP} with \texttt{FedGATE} is strictly smaller than the vanilla \texttt{FedGATE}, when $T_1\leq\dots\leq T_N$. This is due to the fact that \texttt{FLANP} consists of $\log(N)$ stages and its runtime is sum of $\log(N)$ terms each less than or equal to $T_N$.

We characterized the order-wise expressions of expected learning wall-clock time for \texttt{FLANP} and \texttt{FedGATE} methods in Propositions \ref{prop2} and \ref{prop3}. We show in the appendix that these expected delays are precisely characterized as follows
\begin{gather} 
    \E[T_{\texttt{FLANP}}] 
    =
    18 \log(6)c^{-1} \kappa s \sigma^2 \left( T_{n_0} + T_{2n_0} + \cdots + T_N \right), \\
    \E[T_{\texttt{FedGATE}}] 
    =
    7.5 c^{-1} \kappa s \sigma^2 \log \left( 5 c^{-1} \Delta_0 N s \right) T_N,\label{eq: expected times}
\end{gather}
where $\Delta_0 \coloneqq L_N(\w_0) - L_N(\w_N^*)$ denotes the initial sub-optimality corresponding to model $\w_0$. These expressions help better understand the intuition behind the speedup gain provided by \texttt{FLANP}. 
To establish a more concrete speedup gain for the straggler-resilient method, we need to consider a computation time model for the federated clients. Random exponential time model has been widely used to capture the computation delay for distributed clusters \citep{lee2017speeding,reisizadeh2019coded}. In the next theorem, we assume that node computation times are independent realizations of an exponential random variable and characterize the speedup gain of the resilient Algorithm \ref{alg: straggler-robust fed} compared to the benchmark \texttt{FedGATE}.

\begin{theorem} \label{thm: gain}
Suppose the clients' computation times are i.i.d. random variables drawn from an exponential distribution with parameter $\lambda$. That is, $T_1,\cdots,T_N \sim \exp(\lambda)$. Then, the speedup gain of the proposed {\normalfont\texttt{FLANP}} Algorithm \ref{alg: straggler-robust fed} compared to the naive federated learning method {\normalfont\texttt{FedGATE}} is 
\begin{align}
    \frac{\E[T_{\normalfont{\texttt{FLANP}}}]}{\E[T_{\normalfont{\texttt{FedGATE}}}]} \leq \ccalO \left( \frac{1}{\log(Ns)} \right). \nonumber
\end{align}
\end{theorem}

The result in Theorem \ref{thm: gain} establishes a $\ccalO(\log(Ns))$ speedup gain for our proposed straggler-resilient method compared to its non-adaptive and straggler-prone benchmark where the clients' computation times are drawn from a random exponential time model.

We have so far considered heterogeneous federated clients with potentially well-spread computation speeds and demonstrated the speedup gain obtained by adaptive node participation method, particularly in Theorem \ref{thm: gain}. However, this method provides provable speedups even for \emph{homogeneous} federated clients with identical speeds, i.e. $T_1=\cdots=T_N$. 
Comparing the expected runtimes in \eqref{eq: expected times} yields that \texttt{FLANP} in Algorithm \ref{alg: straggler-robust fed} slashes the expected wall-clock time of \texttt{FedGATE} by a factor $\log (Ns)/\log (N)$. This observation demonstrates that the adaptive node participation approach results in two different speedup gains; ($i$) leveraging faster nodes to speedup the learning, and ($ii$) adaptively increase the effective sample size by participating more clients.

\begin{remark}
The result in this section shows that the proposed adaptive node participation framework {\normalfont\texttt{FLANP}} is able to make the {\normalfont\texttt{FedGATE}} algorithm robust against stragglers and reduce its runtime when we face a device heterogeneous setting. We would like to reiterate that {\normalfont\texttt{FLANP}} is a meta-procedure that can be used for any  federated learning solver other than {\normalfont\texttt{FedGATE}} to make it resilient against straggling nodes in the network.  
\end{remark}

\section{Numerical Experiments}

We conduct various numerical experiments for convex and nonconvex risks and evaluate the performance of the proposed  method versus other  benchmarks.

\textbf{Benchmarks.} Bellow is a brief description for multiple federated learning benchmarks that we use to compare with the proposed \texttt{FLANP} in Algorithm \ref{alg: straggler-robust fed}. Note that in all these benchmarks all the available $N$ nodes participate in the training process.  
\vspace{-.3cm}
\begin{itemize}
    \item \texttt{FedAvg} \citep{mcmahan2017communication}. Nodes update their local model using a simple SGD rule for $\tau$ local iterations before uploading to the server.
    \item \texttt{FedGATE} \citep{haddadpour2020federated}. This is the subroutine used in Algorithm \ref{alg: straggler-robust fed}. Here we consider it as a benchmark running with all the available $N$ nodes with model update rule similar to the subroutine in Algorithm \ref{alg: straggler-robust fed}.
    \item \texttt{FedNova} \citep{wang2020tackling}. In each round, each node $i$ updates  its local model for $\tau_i$ iterations where $\tau_i$s vary across the nodes. To mitigate the heterogeneity in $\tau_i$s, the parameter server aggregates normalized updates (w.r.t. $\tau_i$) from the clients and updates the global model.
\end{itemize}

We numerically evaluate the performance of \texttt{FLANP} with such benchmarks both with respect to communication rounds and wall-clock time. We examine \texttt{FLANP} against the benchmarks under both full and partial node participation scenarios and highlight its practicality.

\subsection{Uniform computation speeds} \label{subsec: uniform speed}

\begin{figure}[t]
\centering
\begin{minipage}{.5\textwidth}
    \centering
    \includegraphics[width=0.47\textwidth]{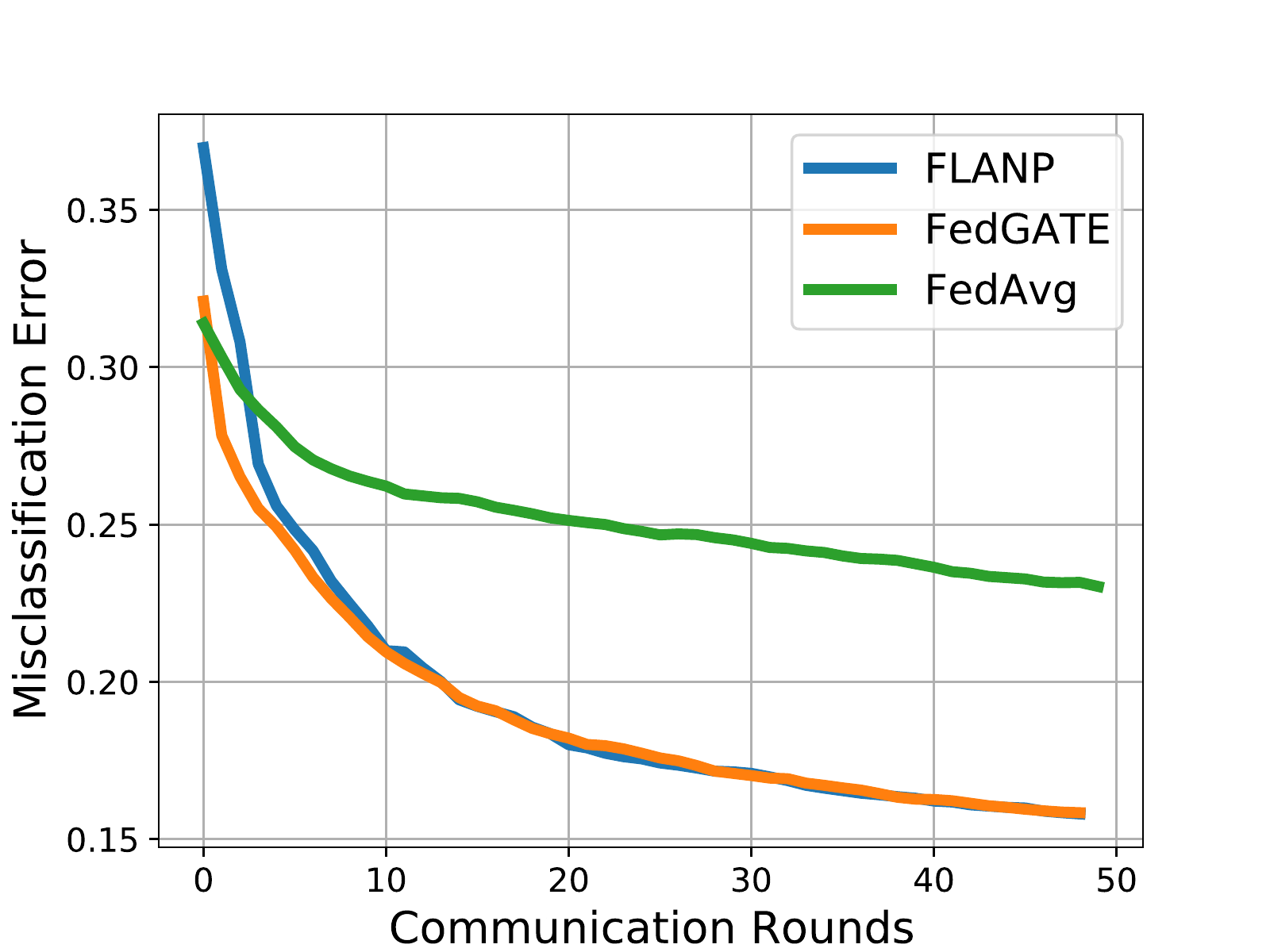}
    \includegraphics[width=0.47\textwidth]{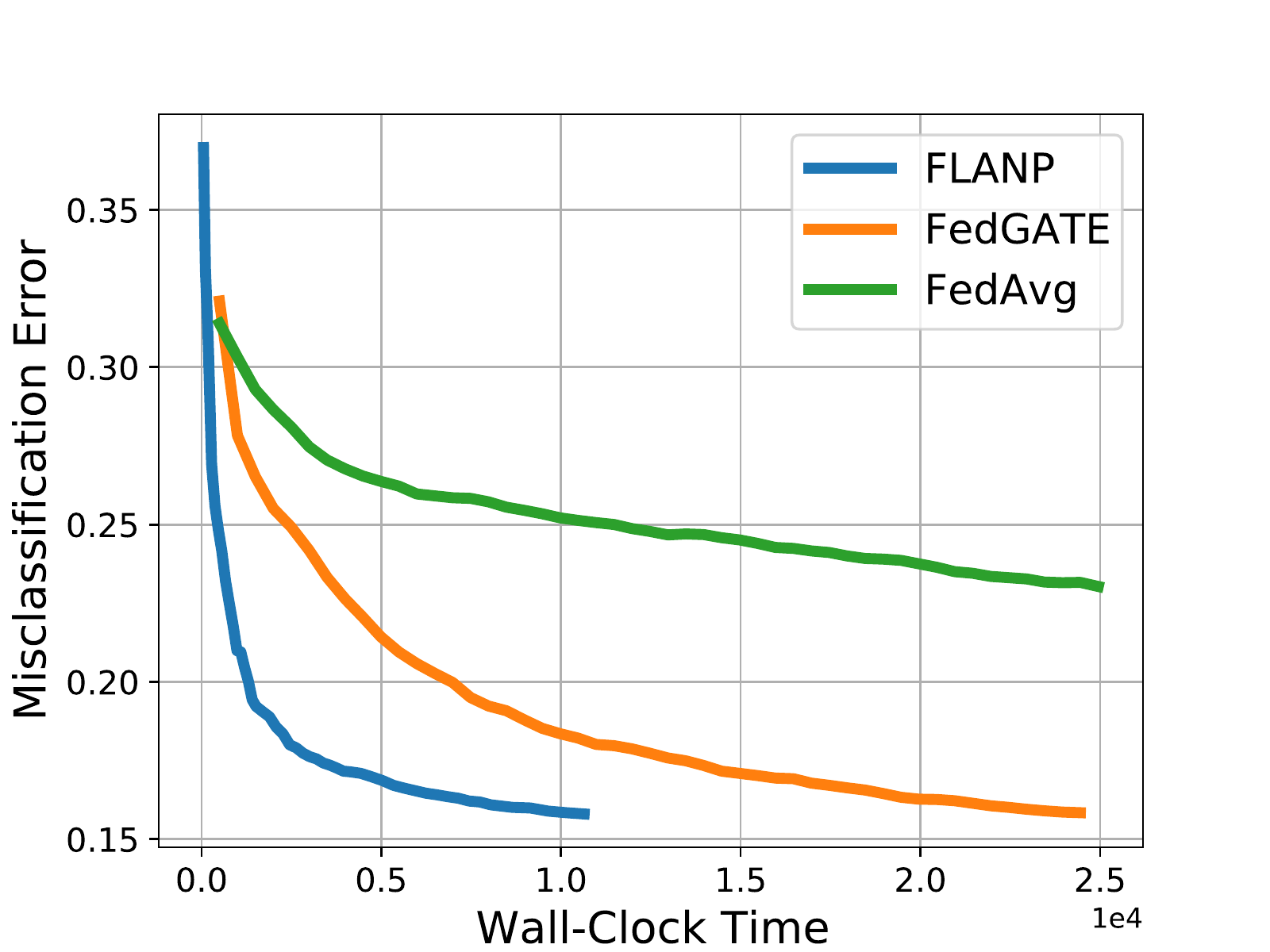}
    \caption{Logistic Regression over MNIST}
    \label{logistic-MNIST}
\end{minipage}%
\begin{minipage}{.5\textwidth}
    \centering
    \includegraphics[width=0.47\textwidth]{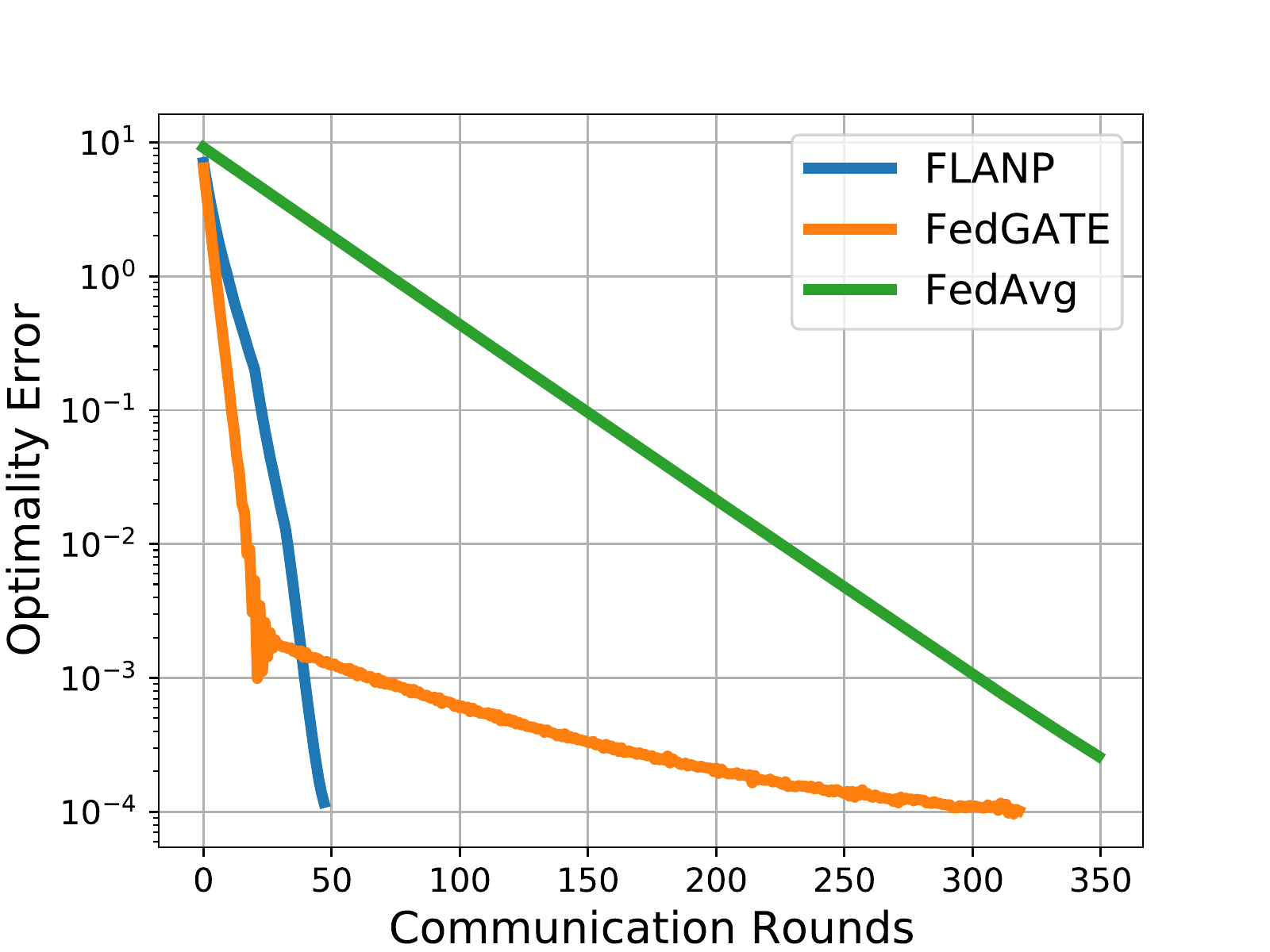}
    \includegraphics[width=0.47\textwidth]{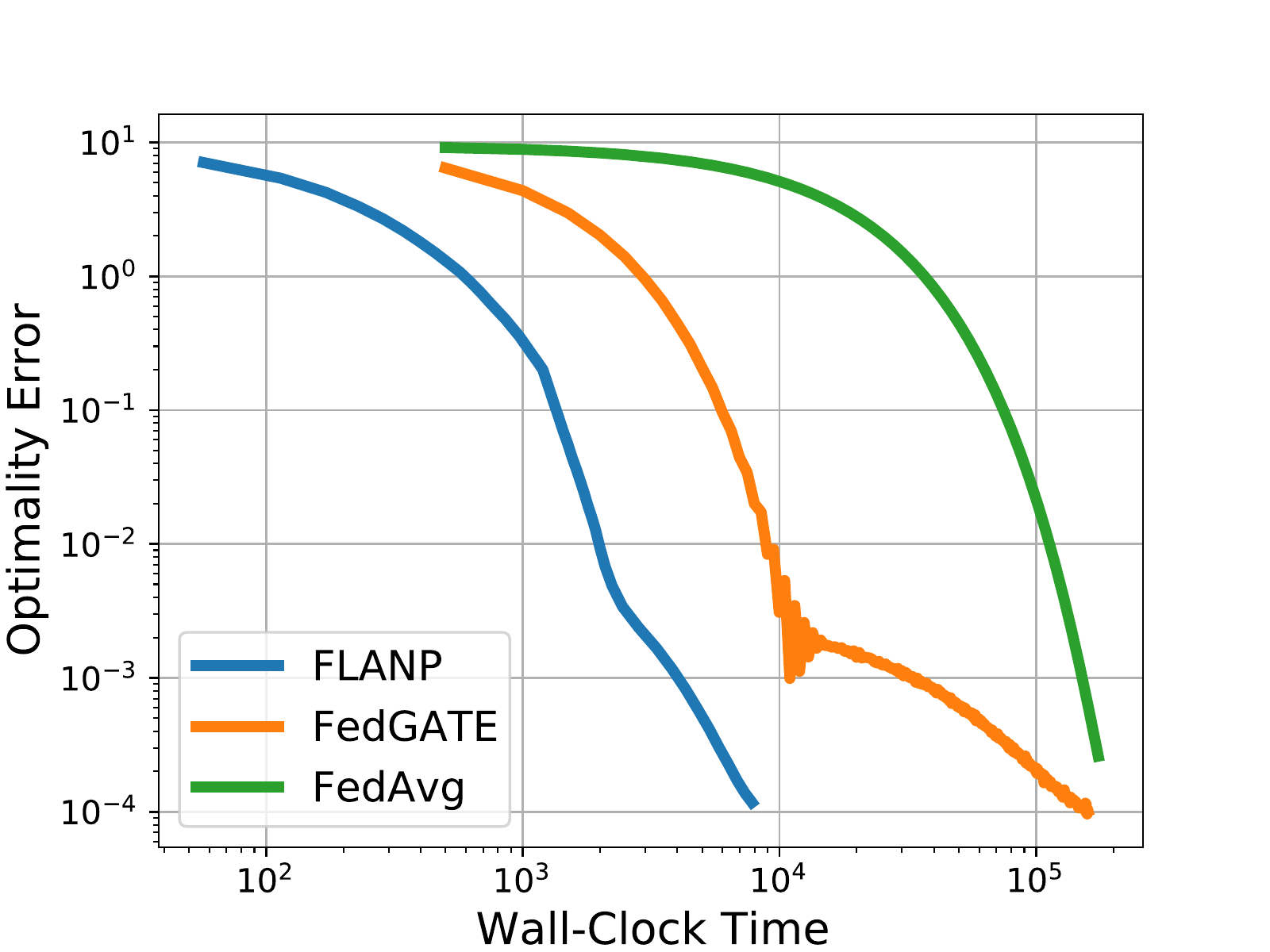}
    \caption{Linear Regression with Synthetic Data}
    \label{linear-Synthetic}
\end{minipage}
\end{figure}

\begin{figure}[t]
\centering
\begin{minipage}{.5\textwidth}
    \centering
    \includegraphics[width=0.47\textwidth]{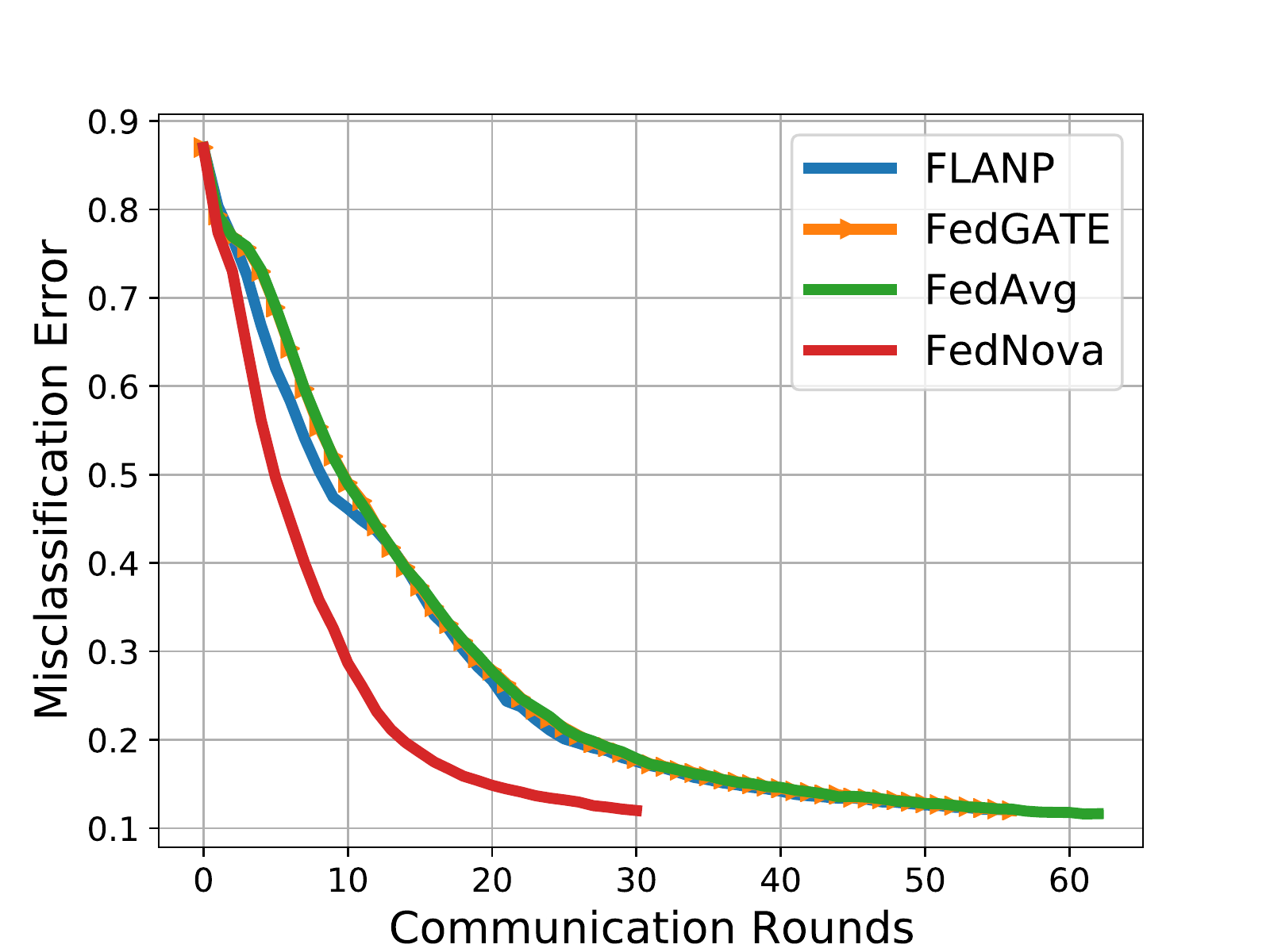}
    \includegraphics[width=0.47\textwidth]{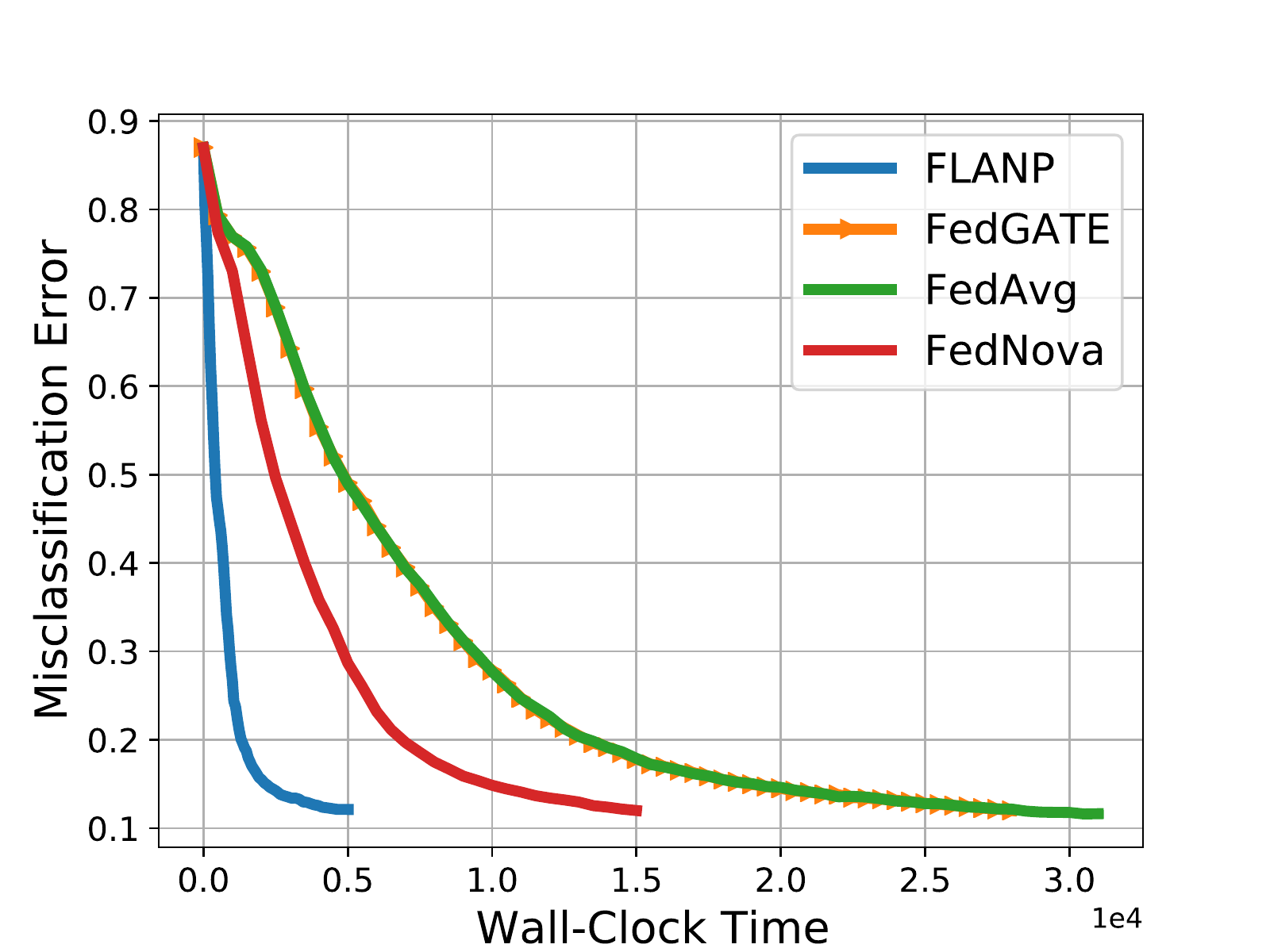}
    \caption{Neural Network Training over MNIST}
    \label{MNIST_nn}
\end{minipage}%
\begin{minipage}{.5\textwidth}
    \centering
    \includegraphics[width=0.47\textwidth]{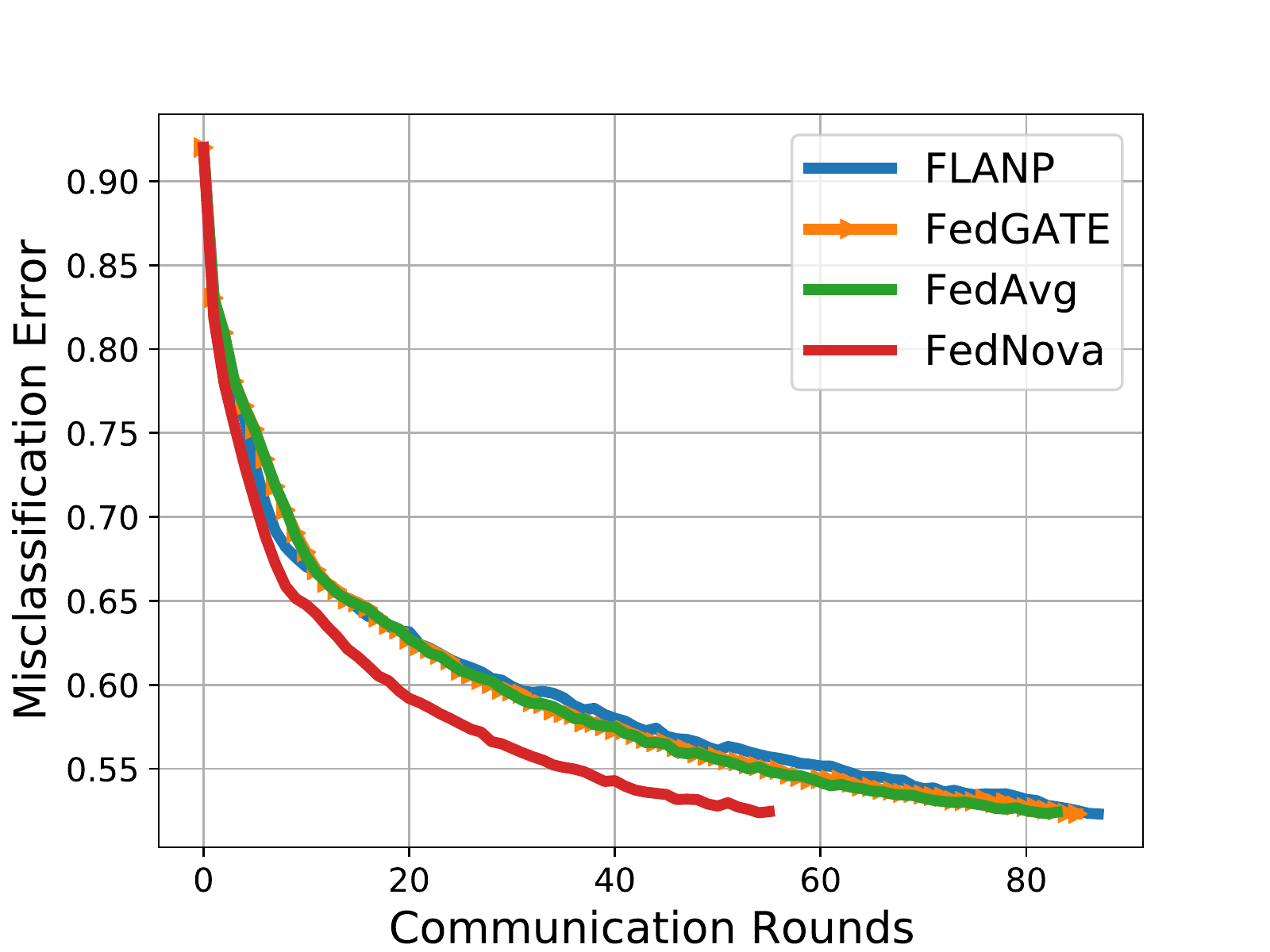}
    \includegraphics[width=0.47\textwidth]{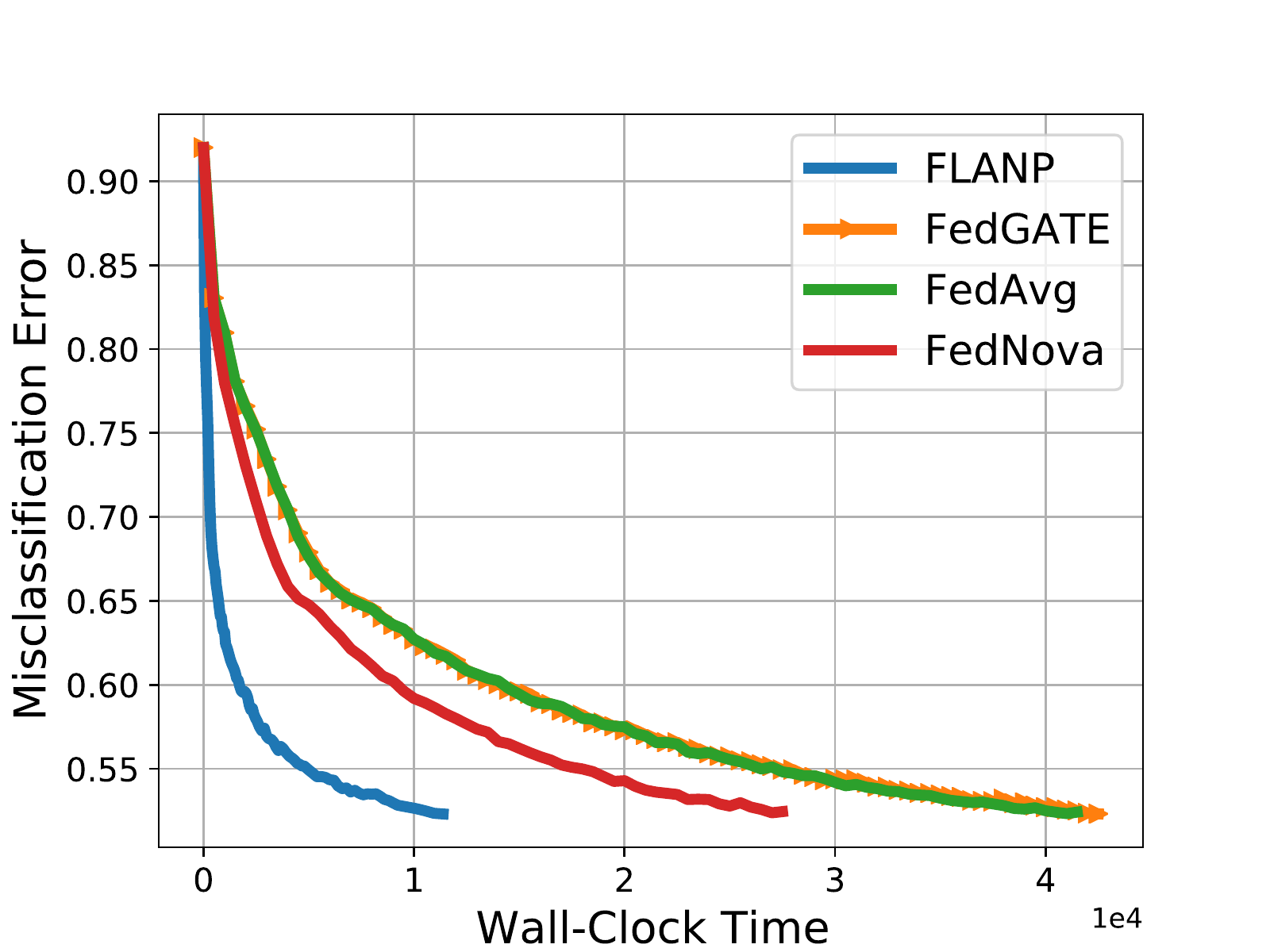}
    \caption{Neural Network Training over CIFAR10}
    \label{CIFAR10_nn}
\end{minipage}
\end{figure}

\textbf{Data and Network.} We use three main datasets for different problems: MNIST ($60,000$ training, $10,000$ test samples), CIFAR10 ($50,000$ training, $10,000$ test samples) and synthetic ($10,000$ samples) datasets. To implement our algorithm, we employ a federated network of $N \in \{20,50,100\}$ heterogeneous clients and in order to model the device heterogeneity, we realize and then fix the computation speed of each node $i$, i.e. $T_i$ from the interval $[50,500]$ uniformly at random.

\textbf{Logistic Regression.} We use the MNIST dataset to learn a  multi-class logistic regression model. In a network of $N=50$ nodes, each client stores $s = 1200$ samples from the MNIST dataset. As demonstrated in Figure \ref{logistic-MNIST} (left), \texttt{FLANP} is slightly outperformed by \texttt{FedGATE} at the initial rounds. This is however expected as \texttt{FLANP} starts with only a fraction of nodes participating which leads to less accurate models. With respect to wall-clock time however, \texttt{FLANP} outperforms both \texttt{FedAvg} and \texttt{FedGATE} benchmarks due to the fact that the initial participating nodes are indeed the fastest ones. As Figure \ref{logistic-MNIST} (right) shows, the adaptive node participation approach leads \texttt{FLANP} to speedup gains of up to $2.1 \times$ compared to \texttt{FedGATE}.

\textbf{Linear Regression.} We train a linear regression model over a synthetic dataset with $10,000$ samples distributed across $N=100$ heterogeneous nodes. Figure \ref{linear-Synthetic} demonstrates convergence of the sub-optimality error of global models $\Vert \w_t-\w^*\Vert$ for three methods \texttt{FLANP}, \texttt{FedGATE} and \texttt{FedAvg}. Similar to Figure \ref{logistic-MNIST}, we observe the same behavior for the optimality gap versus communication round and wall-clock time in Figure \ref{linear-Synthetic} where \texttt{FLANP} manages to speed up the training time by $10 \times$ compared to non-adaptive \texttt{FedGATE}.

\textbf{Neural Network Training.} We train a fully connected neural network with two hidden layers with $128$ and $64$ neurons and compare with three other benchmarks including \texttt{FedNova} which is stragglers-resilient federated algorithm. We conduct two sets of experiments over MNIST and CIFAR10 datasets on a network of $N=20$ clients. Figures \ref{MNIST_nn} and \ref{CIFAR10_nn} demonstrate that \texttt{FLANP} significantly accelerates the training by up to $3\times$ and $4\times$ compared to \texttt{FedNova}. We note that in order to have a fair comparison, the stepsize in all methods is set to $\eta = 0.05$ for MNIST and $\eta = 0.02$ for CIFAR10 and $\gamma=1$.

\subsection{Random exponential computation speeds}

\begin{figure}[h]
\centering
    \includegraphics[width=0.24\textwidth]{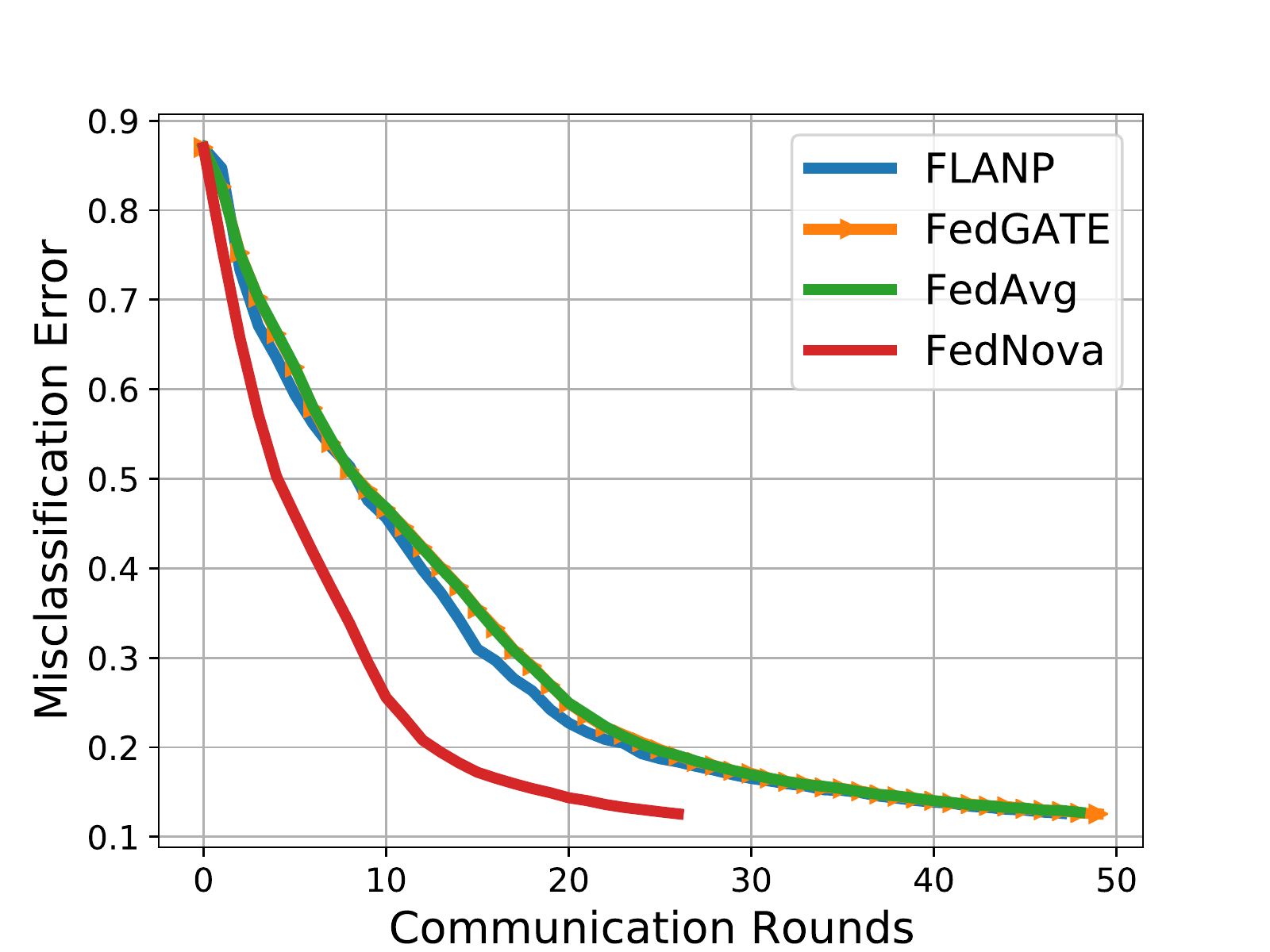}
    \hspace{.5cm}
    \includegraphics[width=0.24\textwidth]{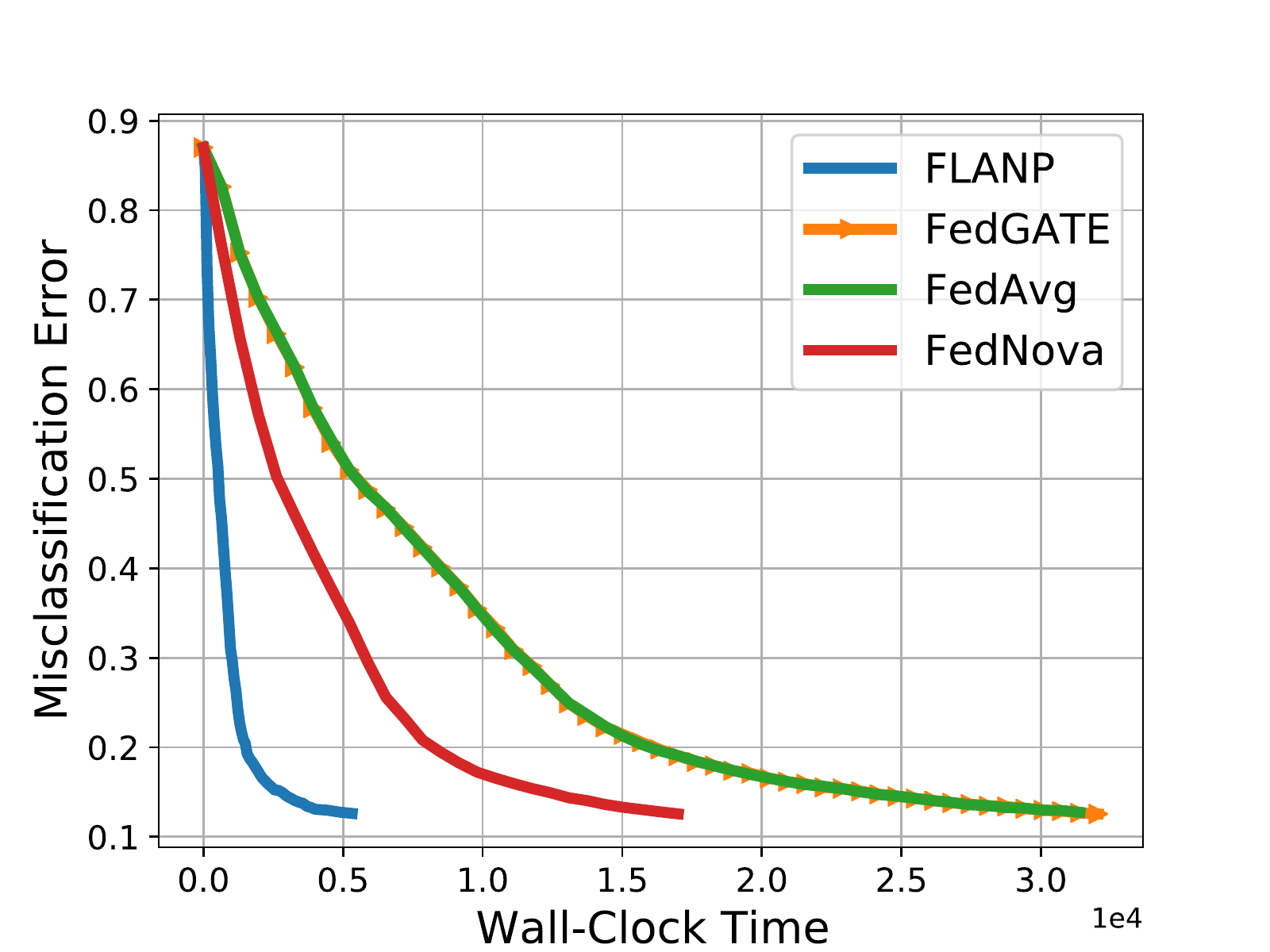}
    \caption{Neural Network training with random exponential speeds on MNIST.}
    \label{fig: exponential MNIST}
\end{figure}

We conduct another set of experiments using the same setup described in Section \ref{subsec: uniform speed}, however, we here pick the clients' computation speed to be i.i.d. random exponential variables which is more consistent with the setup of theorem \ref{thm: gain}. We train a fully connected neural network with two hidden layers with $128$ and $64$ neurons on MNIST dataset and compare with the same benchmarks \texttt{FedAvg}, \texttt{FedGATE} and \texttt{FedNova} as demonstrate in Figure \ref{fig: exponential MNIST}.

\subsection{Comparison with partial node participation methods}

\begin{figure}[h]
\centering
\begin{subfigure}{.3\textwidth}
  \centering
  \includegraphics[width=.84\textwidth]{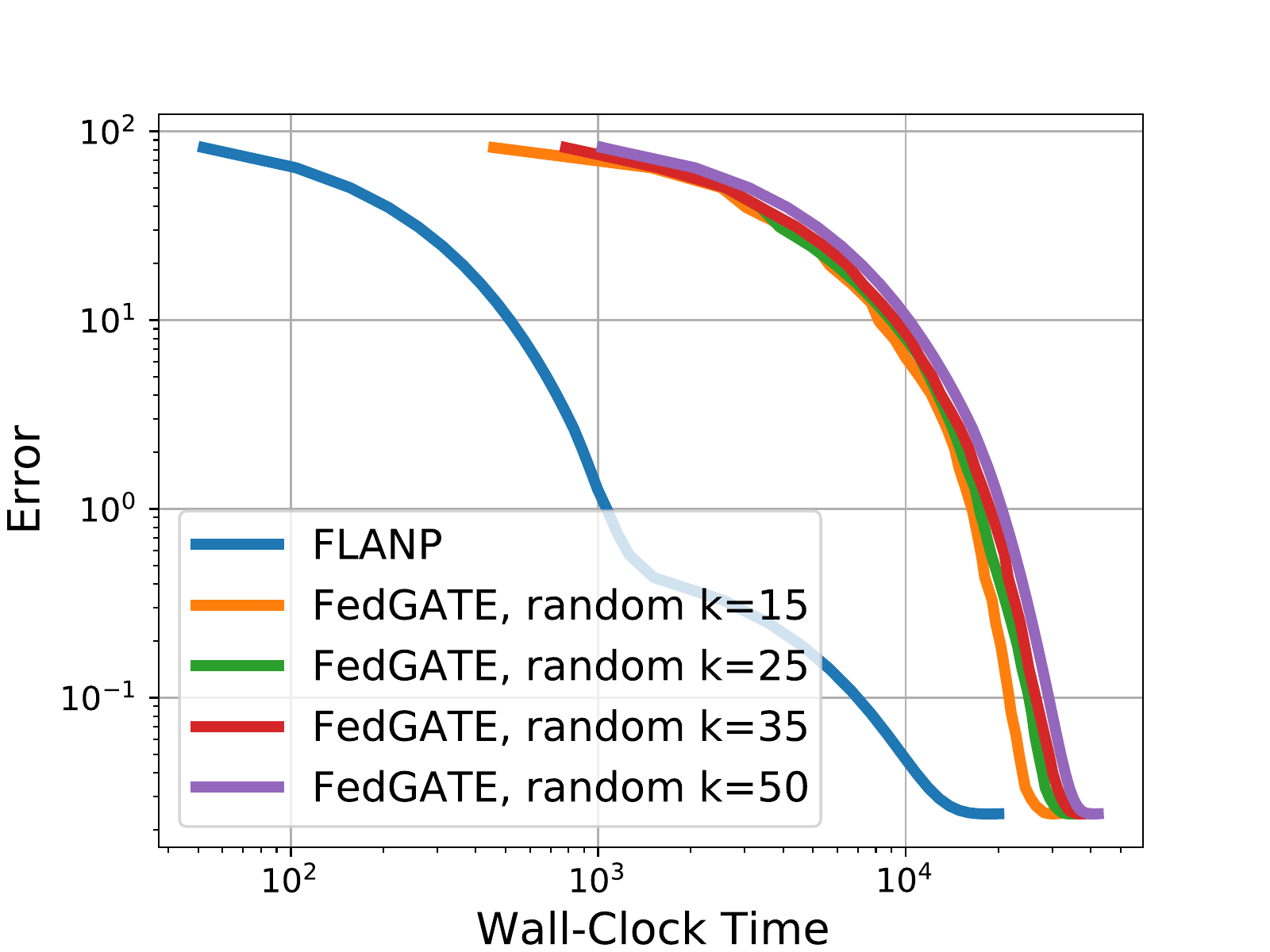}
  \caption{$k$ nodes are randomly picked.}
  \label{fig:partial_random}
\end{subfigure}%
\begin{subfigure}{.3\textwidth}
  \centering
  \includegraphics[width=.84\textwidth]{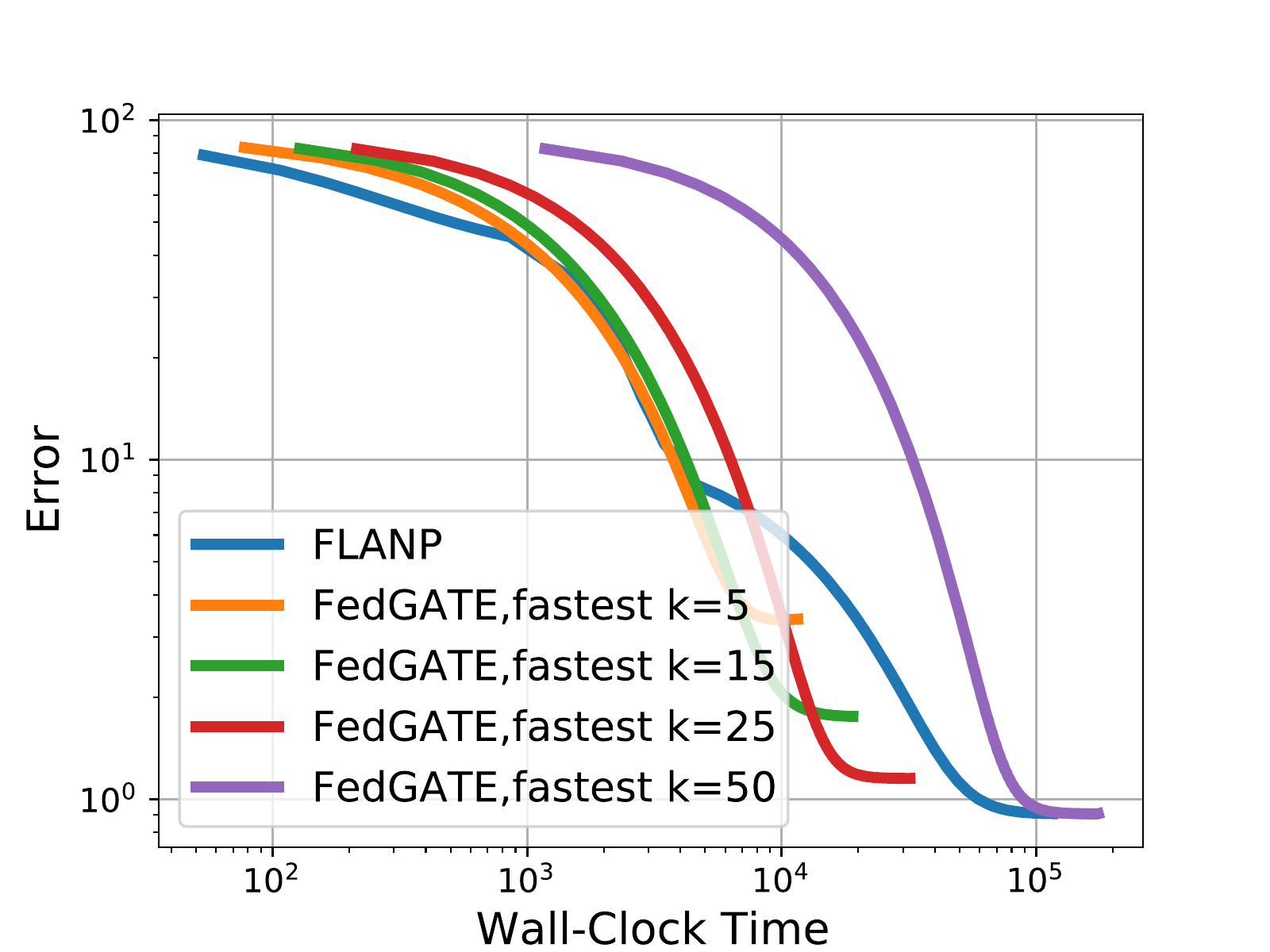}
  \caption{$k$ fastest nodes are picked.}
  \label{fig:partial_fastest}
\end{subfigure}
\caption{\texttt{FLANP} vs. \texttt{FedGATE} with partial node participation.}
\label{fig:partial}
\end{figure}

In this section, we have been comparing the \texttt{FLANP} method with federated benchmarks in which all of the available nodes participate in training in every round. To demonstrate the resiliency of \texttt{FLANP} to partial node participation methods, we consider two different scenarios. First, we compare the wall-clock time of a neural network training of \texttt{FLANP} with partial node participation \texttt{FedGATE} in which only $k$ out of $N=50$ nodes are randomly picked and participate in each round. As demonstrated in Figure \ref{fig:partial}(\subref{fig:partial_random}), \texttt{FLANP} is significantly faster than \texttt{FedGATE} with partial node participation. Second, we consider the case that the $k$ participating nodes are not randomly picked; rather are the fastest clients. As shown in Figure \ref{fig:partial}(\subref{fig:partial_fastest}), although partial participation methods with $k$ fastest nodes begin to outperform \texttt{FLANP} towards the end of the training, they suffer from higher training error saturation as the data samples of \emph{only} $k$ nodes contribute in the learned model and hence the final model is significantly inaccurate.

\subsection{Effect of $N$ and $s$ and the practicality of \texttt{FLANP}}

\begin{figure}[h]
\centering
    \includegraphics[width=0.24\textwidth]{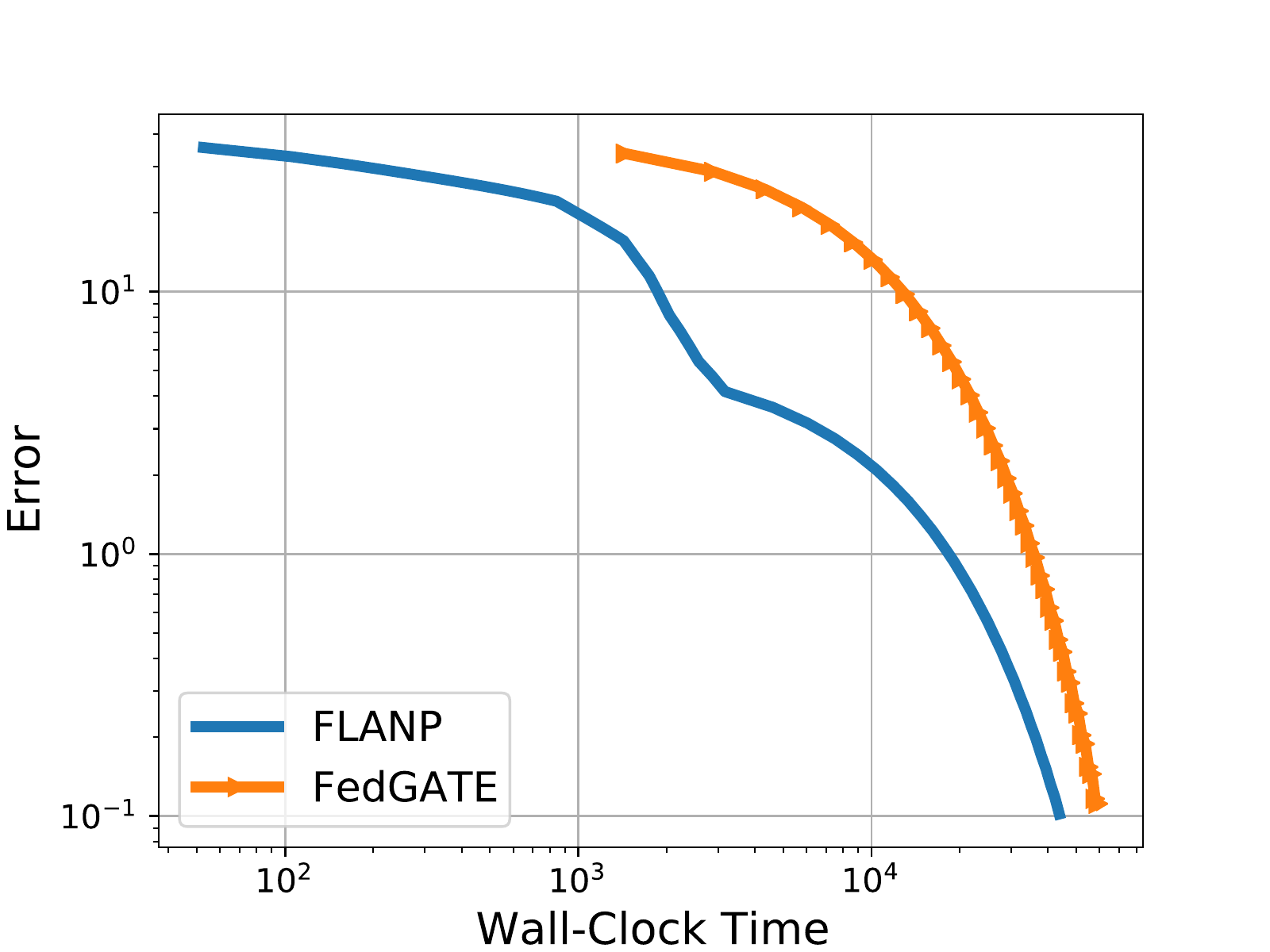}
    \includegraphics[width=0.24\textwidth]{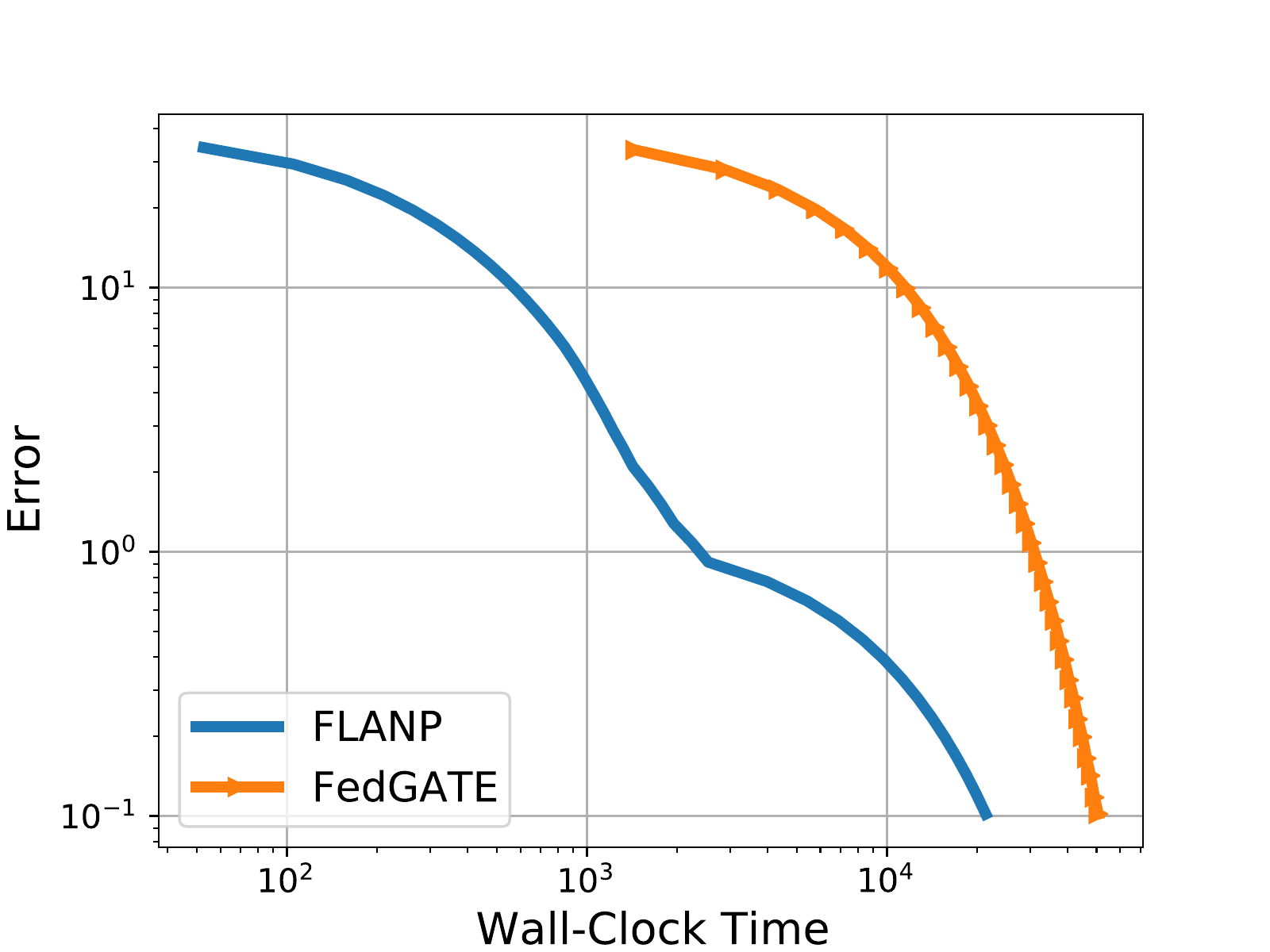}
    \includegraphics[width=0.24\textwidth]{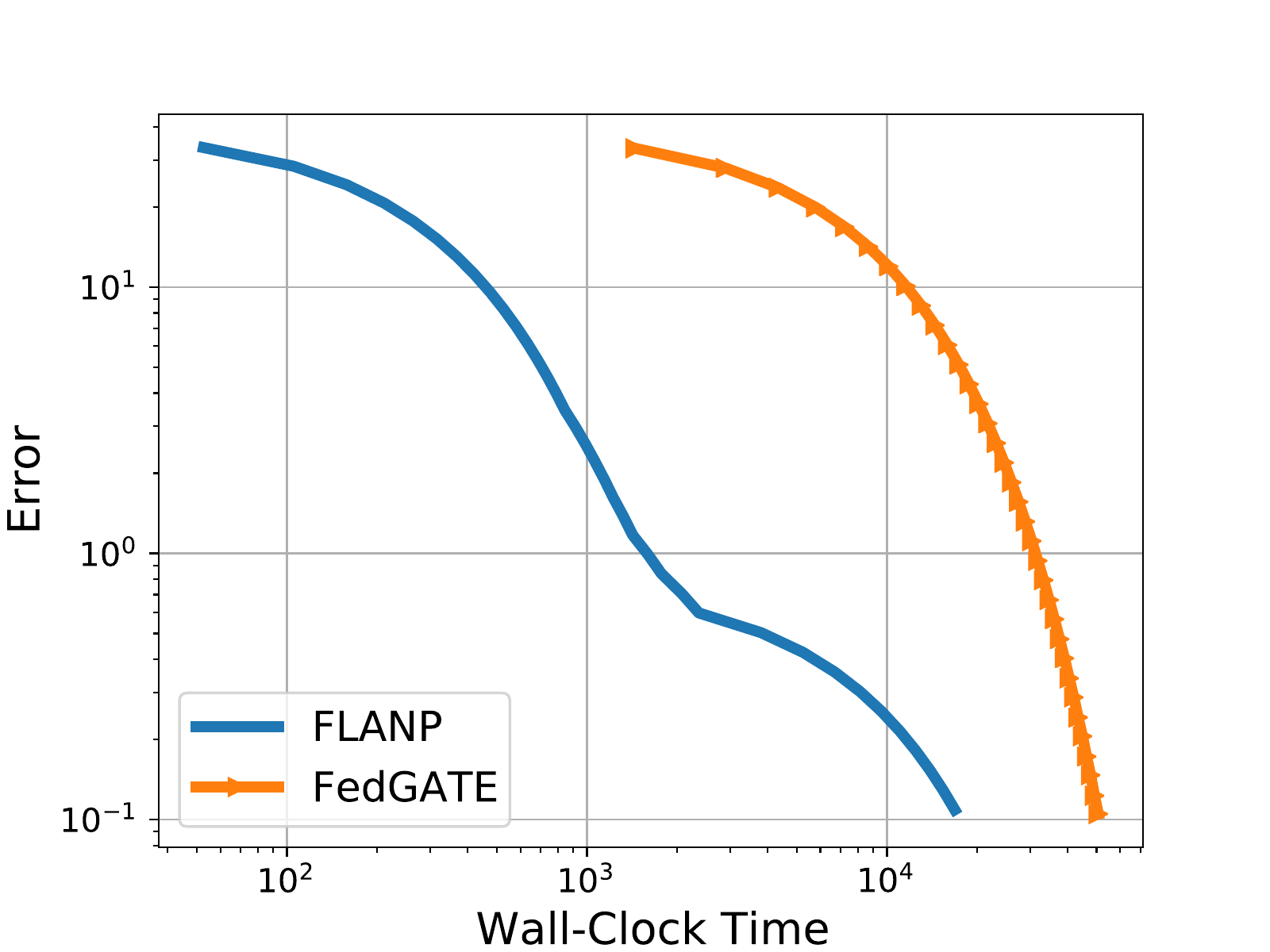}
    \caption{Sub-optimality error vs. wall-clock time for $N=50$ and  $s=20$ (left), $s=200$ (middle), $s=2000$ (right). }
    \label{fig: change s}
\end{figure}

\begin{figure}[h]
\centering
    \includegraphics[width=0.24\textwidth]{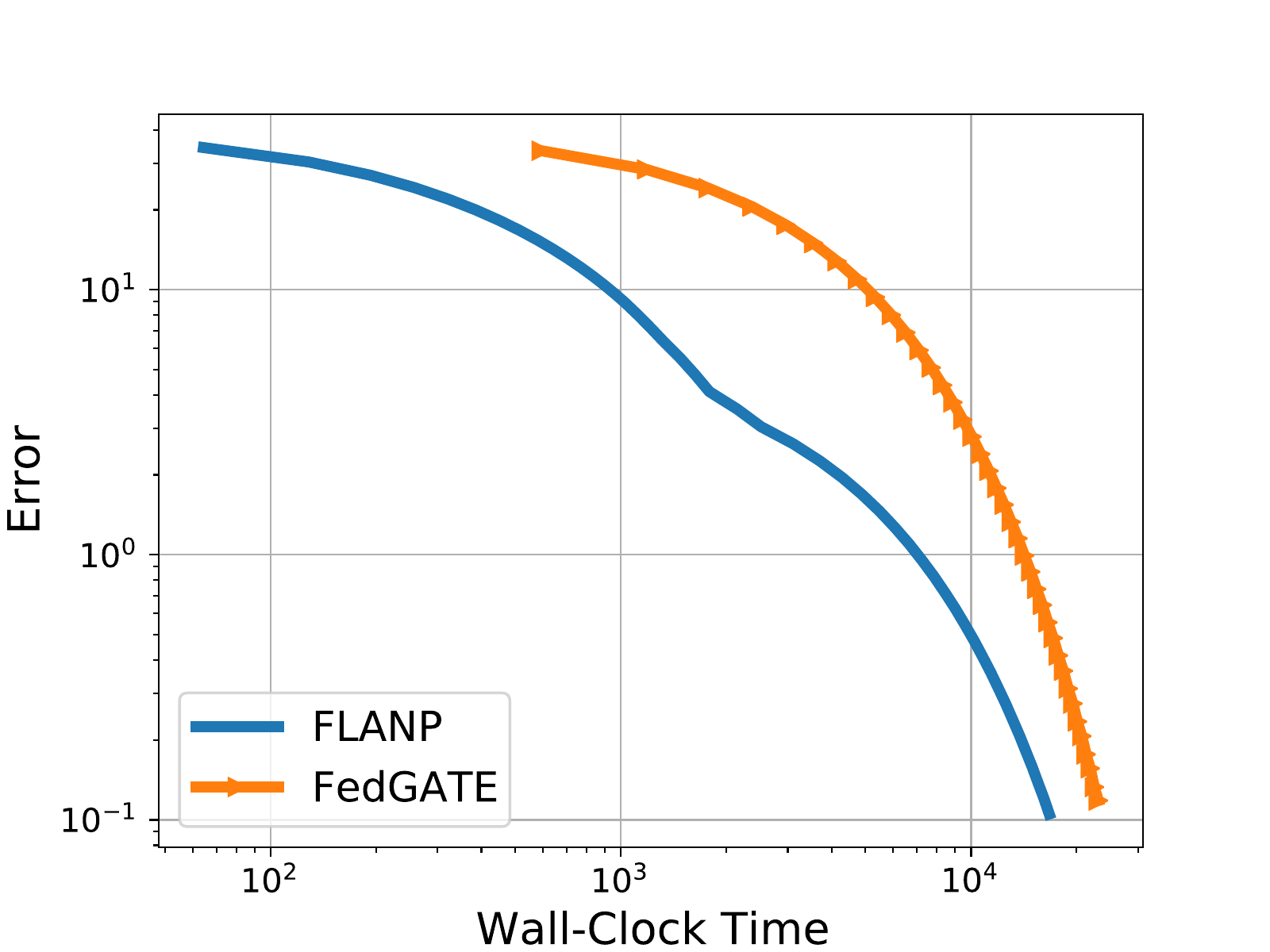}
    \includegraphics[width=0.24\textwidth]{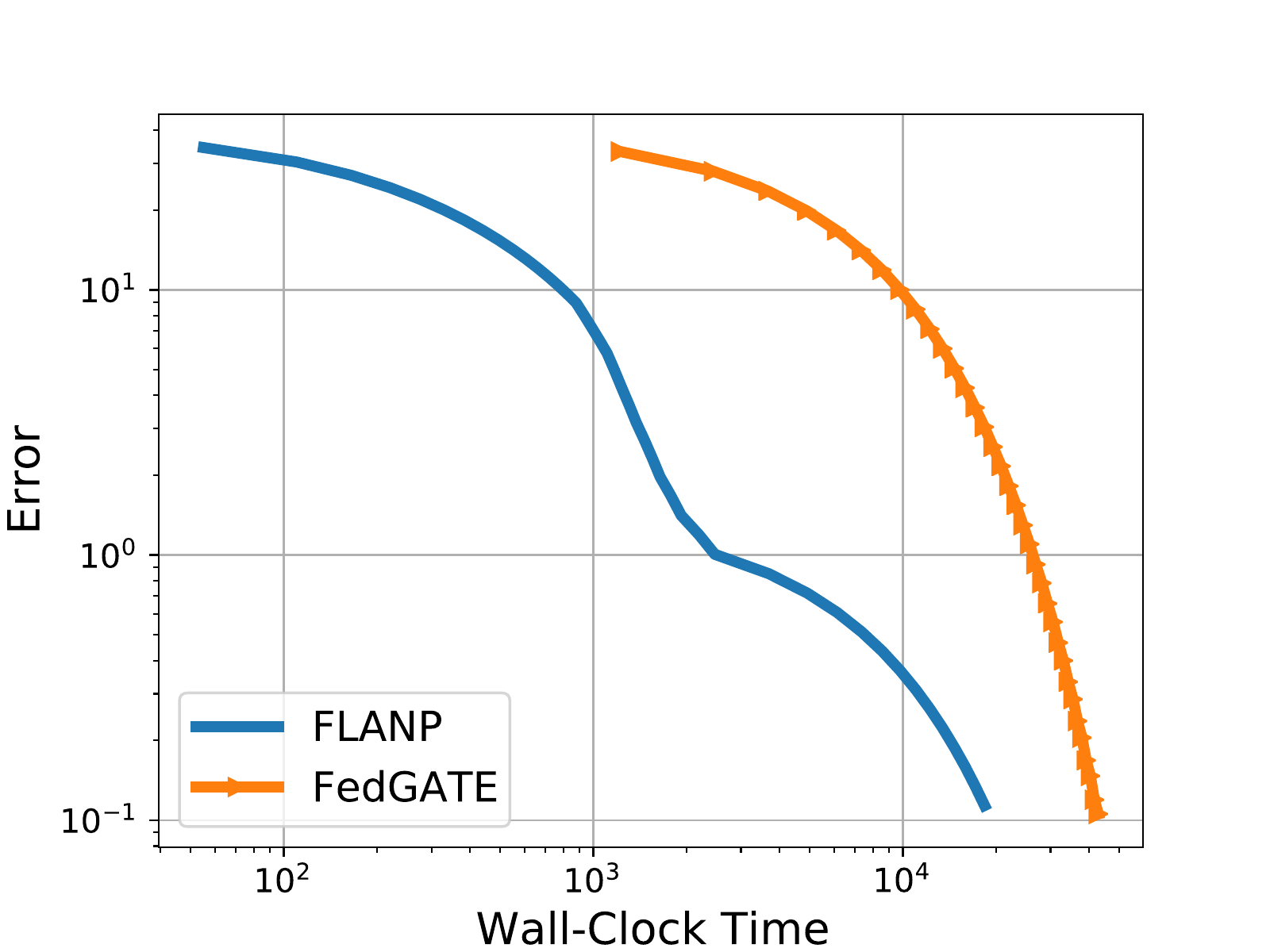}
    \includegraphics[width=0.24\textwidth]{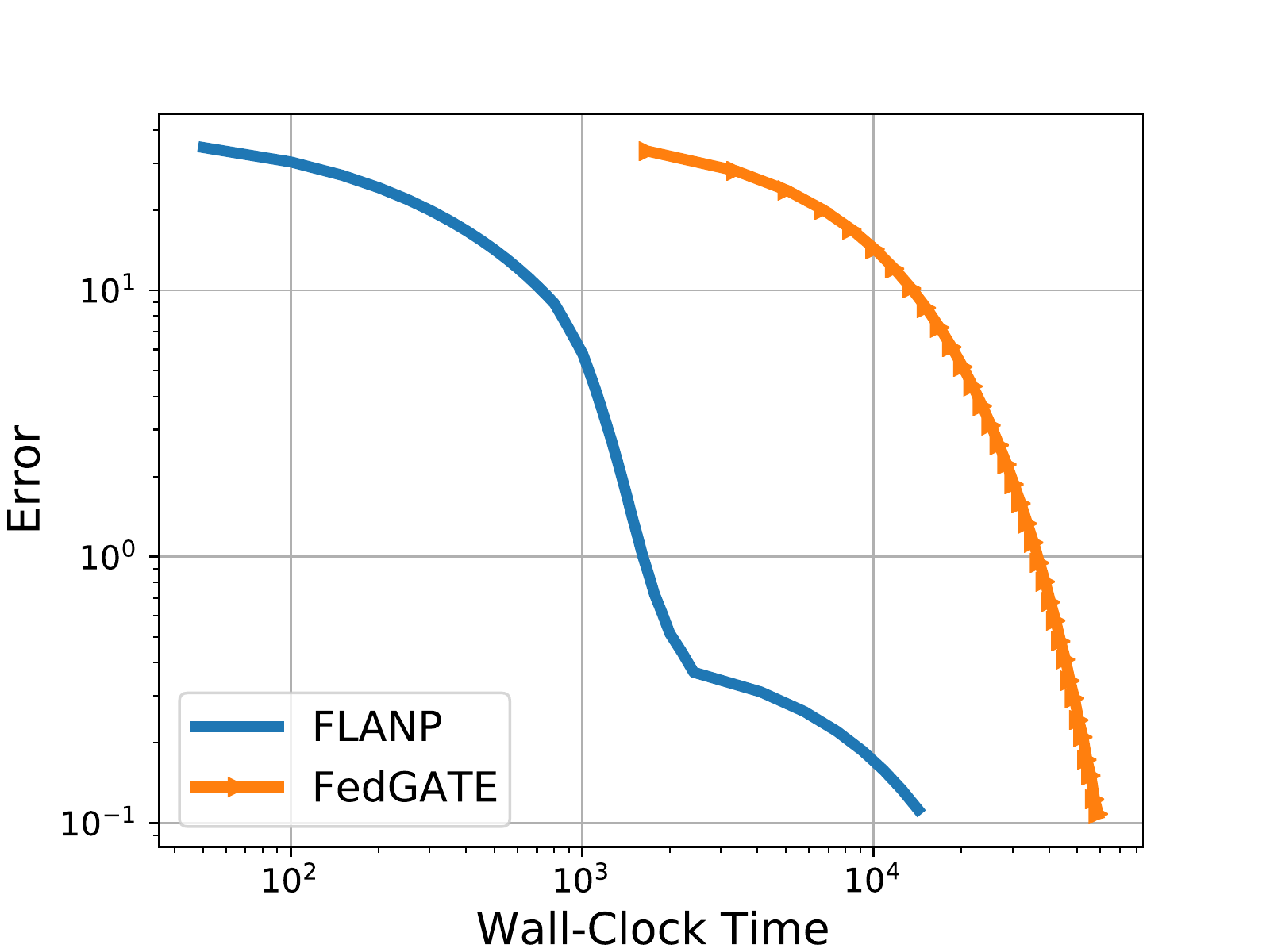}
    \caption{Sub-optimality error vs. wall-clock time for $s=100$ and  $N=10$ (left), $N=100$ (middle), $N=1000$ (right). }
    \label{fig: change N}
\end{figure}

In this section, we particularly focus on the setting described in Theorem \ref{thm: gain} and examine the effect of parameters $N$ and $s$ on the $\ccalO(\log(Ns))$ gain established in Theorem \ref{thm: gain}. We set the nodes' computation speeds to be i.i.d. random exponential variables and run two federated algorithms, \texttt{FLANP} and \texttt{FedGATE}. Figure \ref{fig: change s} demonstrates the running time of the two algorithms for a linear regression problem with synthetic data. In these plots, the number of clients is fixed to be $N=50$ while we increase the number of data samples per node $s \in \{20,200,2000\}$. Moreover, Table \ref{table: change s} shows the individual running time for \texttt{FLANP} and \texttt{FedGATE} methods and the respective ratio corresponding to each case $s \in \{20,200,2000\}$. Next, we conduct experiments with the same setup to examine the effect of $N$ on the speedup gain $\ccalO(\log(Ns))$ established in Theorem \ref{thm: gain}. Figure \ref{fig: change N} and Table \ref{table: change N} demonstrate the running times and the speedup gains for \texttt{FLANP} and \texttt{FedGATE}. These sets of experiments further illustrate that increasing either the number of clients (i.e. larger device heterogeneity)  or the number of data samples per node results in faster training time for the proposed \texttt{FLANP} method.

\begin{table}[h!]
\parbox{.5\linewidth}{
\centering
{\footnotesize
\begin{tabular}{ || c c c c || }
\hline
$s$ & $T_\texttt{FLANP}$ & $T_\texttt{FedGATE}$ & $T_\texttt{FLANP}/T_\texttt{FedGATE}$ \\ 
\hline \hline
$20$ & $4.51 \times 10^4$ & $6.07 \times 10^4$ & $0.74$ \\[.2ex]
$200$ & $2.27 \times 10^4$ & $5.21 \times 10^4$ & $0.43$ \\[.2ex]
$2000$ & $1.82 \times 10^4$ & $5.21 \times 10^4$ & $0.35$ \\ [.1ex]
\hline
\end{tabular}}
\caption{Table to test captions and labels}
\label {table: change s}
}
\hfill
\parbox{.5\linewidth}{
\centering
{\footnotesize
\begin{tabular}{ || c c c c || }
\hline
$N$ & $T_\texttt{FLANP}$ & $T_\texttt{FedGATE}$ & $T_\texttt{FLANP}/T_\texttt{FedGATE}$ \\ 
\hline \hline
$10$ & $1.73 \times 10^4$ & $2.37 \times 10^4$ & $0.73$ \\[.2ex]
$100$ & $1.95 \times 10^4$ & $4.40 \times 10^4$ & $0.44$ \\[.2ex]
$1000$ & $1.59 \times 10^4$ & $6.07 \times 10^4$ & $0.26$ \\ [.1ex]
\hline
\end{tabular}}
\caption{Table to test captions and labels}
\label {table: change N}
}
\end{table}

\setlength{\columnsep}{6.5pt}
\begin{wrapfigure}{r}{0.25\textwidth}
\centering
\vspace{-.2cm}
    \includegraphics[width=0.25\textwidth]{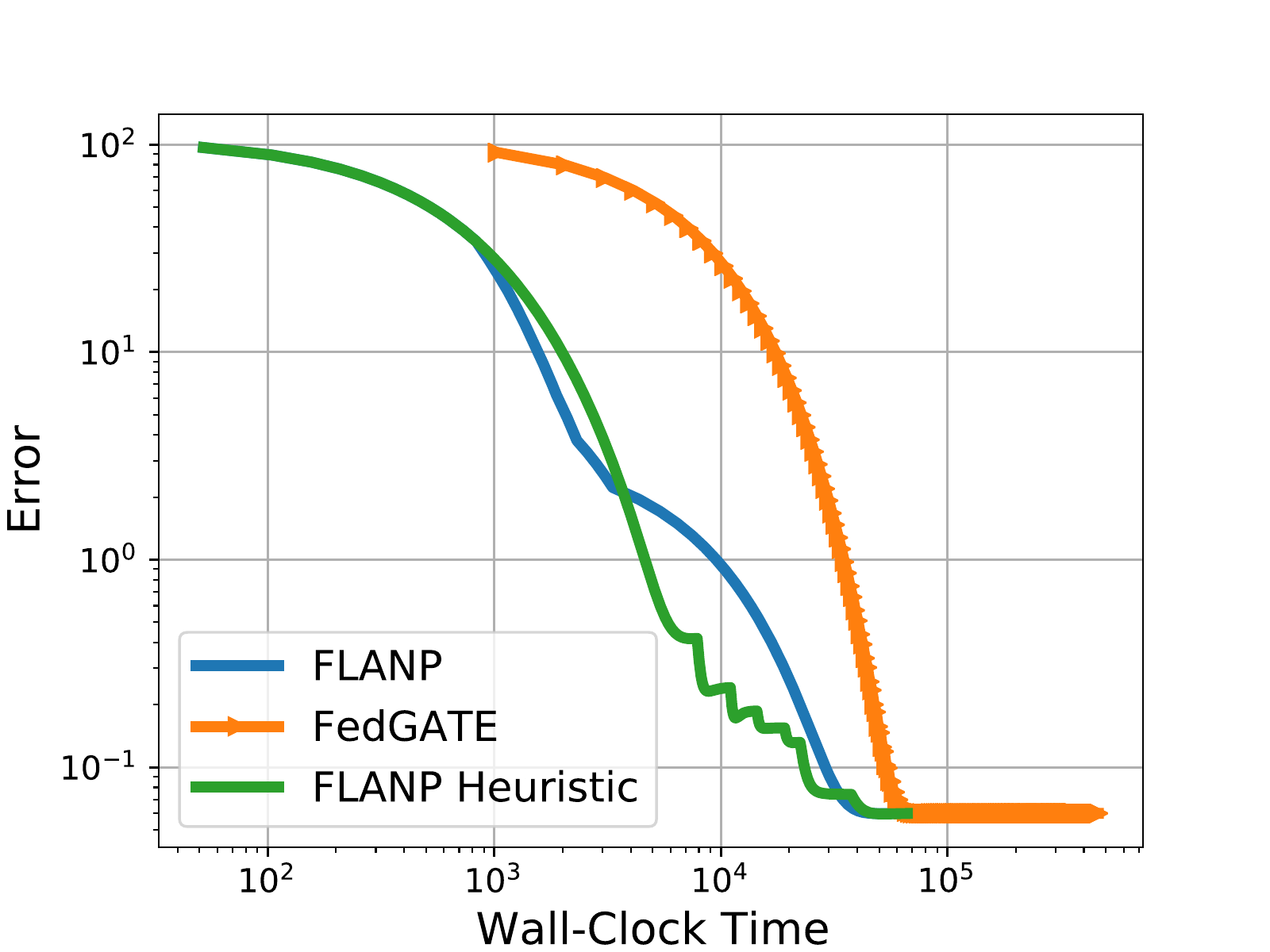}
    \caption{\texttt{FLANP} with heuristic parameter tuning.}
\label{fig:heuristic}
\vspace{-.5cm}
\end{wrapfigure}
Lastly, we highlight a point regarding the knowledge of constant parameters $\mu, c, V_{ns}$ in Algorithm \ref{alg: straggler-robust fed} and the practicality of \texttt{FLANP}. Although such parameters need to be known in order to run \texttt{FLANP}, we note that from the practical point of view, there are several heuristic approaches to handle such issue. We conducted an experiment in which none of the constants are assumed to be known; rather, we heuristically tune the threshold for each phase transition (i.e. doubling the nodes) by monitoring the norm of the global gradient and successively halving the threshold. As shown in Figure \ref{fig:heuristic}, the performance of such heuristic methods is indeed close to \texttt{FLANP} which highlights its practicality.

\clearpage

\bibliography{ref,refs}

\begin{thebibliography}{}

\bibitem[Bartlett et~al., 2006]{bartlett2006convexity}
Bartlett, P.~L., Jordan, M.~I., and McAuliffe, J.~D. (2006).
\newblock Convexity, classification, and risk bounds.
\newblock {\em Journal of the American Statistical Association},
  101(473):138--156.

\bibitem[Bayoumi et~al., 2020]{bayoumi2020tighter}
Bayoumi, A. K.~R., Mishchenko, K., and Richtarik, P. (2020).
\newblock Tighter theory for local sgd on identical and heterogeneous data.
\newblock In {\em International Conference on Artificial Intelligence and
  Statistics}, pages 4519--4529.

\bibitem[Bousquet, 2002]{bousquet2002concentration}
Bousquet, O. (2002).
\newblock {\em Concentration inequalities and empirical processes theory
  applied to the analysis of learning algorithms}.
\newblock PhD thesis, {\'E}cole Polytechnique: Department of Applied
  Mathematics Paris, France.

\bibitem[Eisen et~al., 2018]{eisen2018large}
Eisen, M., Mokhtari, A., and Ribeiro, A. (2018).
\newblock Large scale empirical risk minimization via truncated adaptive
  {Newton} method.
\newblock In {\em AISTATS}.

\bibitem[Frostig et~al., 2015]{frostig2015competing}
Frostig, R., Ge, R., Kakade, S.~M., and Sidford, A. (2015).
\newblock Competing with the empirical risk minimizer in a single pass.
\newblock In {\em Conference on learning theory}, pages 728--763.

\bibitem[Haddadpour et~al., 2019]{haddadpour2019local}
Haddadpour, F., Kamani, M.~M., Mahdavi, M., and Cadambe, V. (2019).
\newblock Local sgd with periodic averaging: Tighter analysis and adaptive
  synchronization.
\newblock In {\em Advances in Neural Information Processing Systems}, pages
  11082--11094.

\bibitem[Haddadpour et~al., 2020]{haddadpour2020federated}
Haddadpour, F., Kamani, M.~M., Mokhtari, A., and Mahdavi, M. (2020).
\newblock Federated learning with compression: Unified analysis and sharp
  guarantees.
\newblock {\em arXiv preprint arXiv:2007.01154}.

\bibitem[Haddadpour and Mahdavi, 2019]{haddadpour2019convergence}
Haddadpour, F. and Mahdavi, M. (2019).
\newblock On the convergence of local descent methods in federated learning.
\newblock {\em arXiv preprint arXiv:1910.14425}.

\bibitem[Huo et~al., 2020]{huo2020faster}
Huo, Z., Yang, Q., Gu, B., Huang, L.~C., et~al. (2020).
\newblock Faster on-device training using new federated momentum algorithm.
\newblock {\em arXiv preprint arXiv:2002.02090}.

\bibitem[Jahani et~al., 2020]{jahani2020efficient}
Jahani, M., He, X., Ma, C., Mokhtari, A., Mudigere, D., Ribeiro, A., and
  Tak{\'a}c, M. (2020).
\newblock Efficient distributed hessian free algorithm for large-scale
  empirical risk minimization via accumulating sample strategy.
\newblock In {\em AISTATS}.

\bibitem[Kairouz et~al., 2019]{kairouz2019advances}
Kairouz, P., McMahan, H.~B., Avent, B., Bellet, A., Bennis, M., Bhagoji, A.~N.,
  Bonawitz, K., Charles, Z., Cormode, G., Cummings, R., et~al. (2019).
\newblock Advances and open problems in federated learning.
\newblock {\em arXiv preprint arXiv:1912.04977}.

\bibitem[Karimireddy et~al., 2019]{karimireddy2019scaffold}
Karimireddy, S.~P., Kale, S., Mohri, M., Reddi, S.~J., Stich, S.~U., and
  Suresh, A.~T. (2019).
\newblock Scaffold: Stochastic controlled averaging for on-device federated
  learning.
\newblock {\em arXiv preprint arXiv:1910.06378}.

\bibitem[Koloskova et~al., 2020]{koloskova2020unified}
Koloskova, A., Loizou, N., Boreiri, S., Jaggi, M., and Stich, S.~U. (2020).
\newblock A unified theory of decentralized sgd with changing topology and
  local updates.
\newblock {\em arXiv preprint arXiv:2003.10422}.

\bibitem[Lee et~al., 2017]{lee2017speeding}
Lee, K., Lam, M., Pedarsani, R., Papailiopoulos, D., and Ramchandran, K.
  (2017).
\newblock Speeding up distributed machine learning using codes.
\newblock {\em IEEE Transactions on Information Theory}, 64(3):1514--1529.

\bibitem[Li et~al., 2019a]{li2019federated}
Li, T., Sahu, A.~K., Talwalkar, A., and Smith, V. (2019a).
\newblock Federated learning: Challenges, methods, and future directions.
\newblock {\em arXiv preprint arXiv:1908.07873}.

\bibitem[Li et~al., 2018]{li2018federated}
Li, T., Sahu, A.~K., Zaheer, M., Sanjabi, M., Talwalkar, A., and Smith, V.
  (2018).
\newblock Federated optimization in heterogeneous networks.
\newblock {\em arXiv preprint arXiv:1812.06127}.

\bibitem[Li et~al., 2019b]{li2019convergence}
Li, X., Huang, K., Yang, W., Wang, S., and Zhang, Z. (2019b).
\newblock On the convergence of fedavg on non-iid data.
\newblock In {\em International Conference on Learning Representations}.

\bibitem[Liang et~al., 2019]{liang2019variance}
Liang, X., Shen, S., Liu, J., Pan, Z., Chen, E., and Cheng, Y. (2019).
\newblock Variance reduced local sgd with lower communication complexity.
\newblock {\em arXiv preprint arXiv:1912.12844}.

\bibitem[Malinovsky et~al., 2020]{malinovsky2020local}
Malinovsky, G., Kovalev, D., Gasanov, E., Condat, L., and Richtarik, P. (2020).
\newblock From local sgd to local fixed point methods for federated learning.
\newblock {\em arXiv preprint arXiv:2004.01442}.

\bibitem[McMahan et~al., 2017]{mcmahan2017communication}
McMahan, B., Moore, E., Ramage, D., Hampson, S., and y~Arcas, B.~A. (2017).
\newblock Communication-efficient learning of deep networks from decentralized
  data.
\newblock In {\em Artificial Intelligence and Statistics}, pages 1273--1282.
  PMLR.

\bibitem[Mishchenko et~al., 2019]{mishchenko2019distributed}
Mishchenko, K., Gorbunov, E., Tak{\'a}{\v{c}}, M., and Richt{\'a}rik, P.
  (2019).
\newblock Distributed learning with compressed gradient differences.
\newblock {\em arXiv preprint arXiv:1901.09269}.

\bibitem[Mohri et~al., 2019]{mohri2019agnostic}
Mohri, M., Sivek, G., and Suresh, A.~T. (2019).
\newblock Agnostic federated learning.
\newblock In {\em International Conference on Machine Learning}, pages
  4615--4625.

\bibitem[Mokhtari et~al., 2016]{mokhtari2016adaptive}
Mokhtari, A., Daneshmand, H., Lucchi, A., Hofmann, T., and Ribeiro, A. (2016).
\newblock Adaptive {Newton} method for empirical risk minimization to
  statistical accuracy.
\newblock In {\em NeurIPS}.

\bibitem[Mokhtari et~al., 2019]{mokhtari2019efficient}
Mokhtari, A., Ozdaglar, A., and Jadbabaie, A. (2019).
\newblock Efficient nonconvex empirical risk minimization via adaptive sample
  size methods.
\newblock In {\em AISTATS}.

\bibitem[Mokhtari and Ribeiro, 2017]{mokhtari2017first}
Mokhtari, A. and Ribeiro, A. (2017).
\newblock First-order adaptive sample size methods to reduce complexity of
  empirical risk minimization.
\newblock In {\em NeurIPS}.

\bibitem[Nishio and Yonetani, 2019]{nishio2019client}
Nishio, T. and Yonetani, R. (2019).
\newblock Client selection for federated learning with heterogeneous resources
  in mobile edge.
\newblock In {\em ICC 2019-2019 IEEE International Conference on Communications
  (ICC)}, pages 1--7. IEEE.

\bibitem[Pathak and Wainwright, 2020]{pathak2020fedsplit}
Pathak, R. and Wainwright, M.~J. (2020).
\newblock Fedsplit: An algorithmic framework for fast federated optimization.
\newblock {\em arXiv preprint arXiv:2005.05238}.

\bibitem[Reddi et~al., 2020]{reddi2020adaptive}
Reddi, S., Charles, Z., Zaheer, M., Garrett, Z., Rush, K., Kone{\v{c}}n{\`y},
  J., Kumar, S., and McMahan, H.~B. (2020).
\newblock Adaptive federated optimization.
\newblock {\em arXiv preprint arXiv:2003.00295}.

\bibitem[Reisizadeh et~al., 2020a]{reisizadeh2020robust}
Reisizadeh, A., Farnia, F., Pedarsani, R., and Jadbabaie, A. (2020a).
\newblock Robust federated learning: The case of affine distribution shifts.
\newblock {\em Advances in Neural Information Processing Systems}, 33.

\bibitem[Reisizadeh et~al., 2020b]{reisizadeh2020fedpaq}
Reisizadeh, A., Mokhtari, A., Hassani, H., Jadbabaie, A., and Pedarsani, R.
  (2020b).
\newblock Fedpaq: A communication-efficient federated learning method with
  periodic averaging and quantization.
\newblock In {\em International Conference on Artificial Intelligence and
  Statistics}, pages 2021--2031.

\bibitem[Reisizadeh et~al., 2019a]{reisizadeh2019coded}
Reisizadeh, A., Prakash, S., Pedarsani, R., and Avestimehr, A.~S. (2019a).
\newblock Coded computation over heterogeneous clusters.
\newblock {\em IEEE Transactions on Information Theory}, 65(7):4227--4242.

\bibitem[Reisizadeh et~al., 2019b]{reisizadeh2019robust}
Reisizadeh, A., Taheri, H., Mokhtari, A., Hassani, H., and Pedarsani, R.
  (2019b).
\newblock Robust and communication-efficient collaborative learning.
\newblock In {\em Advances in Neural Information Processing Systems}, pages
  8388--8399.

\bibitem[Stich, 2019]{stich2019local}
Stich, S.~U. (2019).
\newblock Local sgd converges fast and communicates little.
\newblock In {\em ICLR 2019 ICLR 2019 International Conference on Learning
  Representations}, number CONF.

\bibitem[Stich and Karimireddy, 2019]{stich2019error}
Stich, S.~U. and Karimireddy, S.~P. (2019).
\newblock The error-feedback framework: Better rates for sgd with delayed
  gradients and compressed communication.
\newblock {\em arXiv preprint arXiv:1909.05350}.

\bibitem[Vapnik, 2013]{vapnik2013nature}
Vapnik, V. (2013).
\newblock {\em The nature of statistical learning theory}.
\newblock Springer science \& business media.

\bibitem[Wang and Joshi, 2018a]{wang2018adaptive}
Wang, J. and Joshi, G. (2018a).
\newblock Adaptive communication strategies to achieve the best error-runtime
  trade-off in local-update sgd.
\newblock {\em arXiv preprint arXiv:1810.08313}.

\bibitem[Wang and Joshi, 2018b]{wang2018cooperative}
Wang, J. and Joshi, G. (2018b).
\newblock Cooperative sgd: A unified framework for the design and analysis of
  communication-efficient sgd algorithms.
\newblock {\em arXiv preprint arXiv:1808.07576}.

\bibitem[Wang et~al., 2020]{wang2020tackling}
Wang, J., Liu, Q., Liang, H., Joshi, G., and Poor, H.~V. (2020).
\newblock Tackling the objective inconsistency problem in heterogeneous
  federated optimization.
\newblock {\em arXiv preprint arXiv:2007.07481}.

\bibitem[Wang et~al., 2019]{wang2019adaptive}
Wang, S., Tuor, T., Salonidis, T., Leung, K.~K., Makaya, C., He, T., and Chan,
  K. (2019).
\newblock Adaptive federated learning in resource constrained edge computing
  systems.
\newblock {\em IEEE Journal on Selected Areas in Communications},
  37(6):1205--1221.

\bibitem[Xie et~al., 2019]{xie2019asynchronous}
Xie, C., Koyejo, S., and Gupta, I. (2019).
\newblock Asynchronous federated optimization.
\newblock {\em arXiv preprint arXiv:1903.03934}.

\bibitem[Zhou and Cong, 2017]{zhou2017convergence}
Zhou, F. and Cong, G. (2017).
\newblock On the convergence properties of a $ k $-step averaging stochastic
  gradient descent algorithm for nonconvex optimization.
\newblock {\em arXiv preprint arXiv:1708.01012}.

\end{thebibliography}

\bibliographystyle{apalike}


\appendix

\section{Proof of Proposition \ref{prop}}

Let us present and prove the following lemma which includes the claim in Proposition \ref{prop}.

\begin{lemma} \label{lemma: 3 claims}
Consider two subsets of nodes $\ccalN_m \subseteq \ccalN_n$ and assume that model $\w_m$ attains the statistical accuracy for the empirical risk associated with nodes in $\ccalN_m$, that is, $\Vert \gr L_m (\w_m) \Vert^2 \leq 2 \mu V_{m s}$ where the loss function $\ell$ is $\mu$-strongly convex. Then the suboptimality of $\w_m$ for risk $L_n$, i.e., $L_n(\w_m) - L_n(\w^*_n)$ is w.h.p. bounded above as follows:
\begin{align} \label{eq: error}
    L_n(\w_m) - L_n(\w^*_n) 
    &\leq
    \frac{2(n - m)}{n} \left( V_{(n-m)s} + V_{ms} \right) + V_{ms}.
\end{align}
Moreover, norms of local and global gradients are upper-bounded w.h.p. as follows:
\begin{gather} \label{eq: global gradient}
    \Vert \gr L_n (\w_m) \Vert^2 
    \leq
    2 \left( \frac{n - m}{n} \right)^2 \left( V^{1/2}_{(n-m)s} + V^{1/2}_{ms} \right)^2 + 4 \mu V_{ms},
\end{gather}
and
\begin{gather} \label{eq: local gradient}
    \Vert \gr L^{i} (\w_m) \Vert^2 
    \leq
    3(2\mu + 1) V_{m s} + 3 V_s.
\end{gather}
\end{lemma}

\emph{Proof.} We begin the proof of Lemma \ref{lemma: 3 claims} by proving the inequality in \eqref{eq: error}. Let us decompose the sub-otpimiality error $L_n(\w_m) - L_n(\w^*_n)$ to four difference terms as follows:
\begin{align} \label{eq: L decompose}
    L_n(\w_m) - L_n(\w^*_n) 
    &=
    L_n(\w_m) - L_m(\w_m)
    +
    L_m(\w_m) - L_m(\w^*_m)  +
    L_m(\w^*_m) - L_m(\w^*_n) 
    +
    L_m(\w^*_n) - L_n(\w^*_n).
\end{align}
From definition of local empirical risks in \eqref{eq:L_n}, the difference of local risks $L_n(\w)$ and $L_m(\w)$ for any $\w$ can be bounded w.h.p. as follows:
\begin{align} \label{eq: abs L}
    L_n(\w) - L_m(\w) 
    &\leq
    \abs{L_n(\w) - L_m(\w)} \\
    &=
    \abs{\frac{1}{n} \sum_{i \in \ccalN_n} L^i (\w) - \frac{1}{m} \sum_{i \in \ccalN_m} L^i (\w)} \\
    &=
    \abs{\frac{1}{n} \sum_{i \in \ccalN_n \setminus \ccalN_m} L^i (\w) - \frac{n - m}{n} \cdot \frac{1}{m} \sum_{i \in \ccalN_m} L^i (\w)} \\
    &= 
    \frac{n - m}{n} \abs{\frac{1}{n-m} \sum_{i \in \ccalN_n \setminus \ccalN_m} L^i (\w) - \frac{1}{m} \sum_{i \in \ccalN_m} L^i (\w)} \\
    &\leq
    \frac{n - m}{n} \abs{\frac{1}{n-m} \sum_{i \in \ccalN_n \setminus \ccalN_m} L^i (\w) - L(\w)}
    +
    \frac{n - m}{n} \abs{\frac{1}{m} \sum_{i \in \ccalN_m} L^i (\w) - L(\w)} \\
    &\leq
    \frac{n - m}{n} \left( V_{(n-m)s} + V_{ms} \right),
\end{align}
where the last inequality is  implied from Assumption \ref{assumption: approx error} when applied to empirical risks $\frac{1}{n-m} \sum_{i \in \ccalN_n \setminus \ccalN_m} L^i (\w)$ and $\frac{1}{m} \sum_{i \in \ccalN_m} L^i (\w)$ with $(n-m)s$ and $ms$ samples, respectfully. We now proceed to bound the next term in \eqref{eq: L decompose}, that is the optimality gap $L_m(\w_m) - L_m(\w^*_m)$. Using the strong convexity assumption in Assumption \ref{assumption loss} and the condition $\Vert \gr L_m (\w_m) \Vert^2 \leq 2 \mu V_{m s}$ assumed to hold in the statement of the lemma, we can write
\begin{align}
    L_m(\w_m) - L_m(\w^*_m)
    \leq
    \frac{1}{2 \mu} \norm{\gr L_m (\w_m) }^2 \leq \frac{2 \mu V_{ms}}{2 \mu} = V_{ms}.
\end{align}
Next, the term $L_m(\w^*_m) - L_m(\w^*_n)$ in \eqref{eq: L decompose} can be simply bounded as $L_m(\w^*_m) - L_m(\w^*_n) \leq 0$, since $\w^*_m$ is the minimizer of $L_m(\w)$. Finally, to bound $L_m(\w^*_n) - L_n(\w^*_n) $ in \eqref{eq: L decompose}, we use the result in \eqref{eq: abs L} which holds for any $\w$ and here we pick $\w = \w^*_n$ to conclude
\begin{align}
    L_m(\w^*_n) - L_n(\w^*_n) 
    &\leq
    \frac{n - m}{n} \left( V_{(n-m)s} + V_{ms} \right).
\end{align}
Putting the upper bounds for the four terms in \eqref{eq: L decompose} together proves inequality \eqref{eq: error} which is the same claim as in Proposition \ref{prop}.

Next we prove inequality \eqref{eq: global gradient} by first noting the following:
\begin{align} \label{eq: gradient bound}
    \norm{\gr L_n (\w_m)}^2 
    \leq
    2 \norm{\gr L_n (\w_m) - \gr L_m (\w_m)}^2 + 2 \norm{\gr L_m (\w_m)}^2.
\end{align}
The first term $\Vert \gr L_n (\w_m) - \gr L_m (\w_m) \Vert$ can be bounded as follows:
\begin{align}
    \norm{\gr L_n (\w_m) - \gr L_m (\w_m)}
    &=
    \norm{\frac{1}{n} \sum_{i \in \ccalN_n} \gr L^i (\w) - \frac{1}{m} \sum_{i \in \ccalN_m} \gr L^i (\w)} \\
    &=
    \norm{\frac{1}{n} \sum_{i \in \ccalN_n \setminus \ccalN_m} \gr L^i (\w) - \frac{n - m}{n} \cdot \frac{1}{m} \sum_{i \in \ccalN_m} \gr L^i (\w)} \\
    &= 
    \frac{n - m}{n} \norm{\frac{1}{n-m} \sum_{i \in \ccalN_n \setminus \ccalN_m} \gr L^i (\w) - \frac{1}{m} \sum_{i \in \ccalN_m} \gr L^i (\w)} \\
    &\leq
    \frac{n - m}{n} \norm{\frac{1}{n-m} \sum_{i \in \ccalN_n \setminus \ccalN_m} \gr L^i (\w) - \gr L(\w)}\\
    &\quad +
    \frac{n - m}{n} \norm{\frac{1}{m} \sum_{i \in \ccalN_m} \gr L^i (\w) - \gr L(\w)} \\
    &\leq
    \frac{n - m}{n} \left( V_{(n-m)s}^{1/2} + V_{ms}^{1/2} \right).
\end{align}
In the last inequality above, we used Assumption \ref{assumption: approx error} to upper-bound the approximation of empirical gradients for $(n-m)s$ and $ms$ samples. Together with \eqref{eq: gradient bound} and the assumption of the lemma, that is $\norm{\gr L_m (\w_m)}^2 \leq 2 \mu V_{ms}$, the claim in \eqref{eq: global gradient} is concluded:
\begin{align}
    \Vert \gr L_n (\w_m) \Vert^2 
    \leq
    2 \left( \frac{n - m}{n} \right)^2 \left( V^{1/2}_{(n-m)s} + V^{1/2}_{ms} \right)^2 + 4 \mu V_{ms}.
\end{align}

Finally, we prove the claim in inequality \eqref{eq: local gradient} by bounding node $i$'s local gradient $\gr L^i(\w_m)$ as follows:
\begin{align}
    \norm{ \gr L^i (\w_m) }^2 
    &\leq 
    3 \norm{\gr L^i (\w_m) - \gr L(\w_m)}^2 
    +
    3 \norm{\gr L_m (\w_m) - \gr L(\w_m)}^2
    +
    3 \norm{\gr L_m(\w_m)}^2 \\
    &\leq
    3 V_s + 3 V_{ms} + 6 \mu V_{ms} \\
    &=
    3(2\mu + 1) V_{m s} + 3 V_s,
\end{align}
where we used Assumption \ref{assumption: approx error} to upper-bound the approximation error of empirical gradients for node $i$ with $s$ samples and $m$ nodes with $ms$ samples.

\section{Proof of Theorem \ref{theorem:1}}

Consider a stage of Algorithm \ref{alg: straggler-robust fed} running with $n$ participating nodes. More precisely, $n$ nodes in $\{1,\cdots,n\}$ begin a sequence of local and global model updates according to \texttt{FedGATE} initialized with $\w_m$ obtained from the previous stage ($n=2m$). After $R_n$ communication rounds each with $\tau_n$ local updates, the final sub-optimality error is upper-bounded as follows: (refer to Algorithm 2 and Theorem E.6 in \cite{haddadpour2020federated} with no quantization)
\begin{align} \label{eq: fedgate}
    \E[L_n(\w) - L_n(\w_n^*)]
    &\leq
    \left( 1 - \frac{1}{3} \mu \eta_n \gamma_n \tau_n \right)^{R_n} \left( L_n(\w_m) - L_n(\w_n^*) \right) \\
    &\quad +
    24 \kappa^3 L \tau_n^2 \eta_n^2 \frac{1}{n} \sum_{i=1}^{n} \norm{\gr L^i (\w_m)}^2
    +
    24 \kappa L \tau_n^2 \eta_n^2 \norm{\gr L_n (\w_m)}^2 \\
    &\quad +
    24 \kappa^2 L^2 \tau_n^2 \eta_n^3 \sigma^2  +
    15 \kappa L^3 \tau_n^3 \eta_n^2 (\eta_n \gamma_n)^2 \frac{\sigma^2}{n} + \frac{L}{2} \eta_n \gamma_n \frac{\sigma^2}{n},
\end{align}
where two stepsizes $\eta_n,\gamma_n$ satisfy the following conditions:
\begin{align} \label{eq: stepsize cond}
    1 - L \eta_n \gamma_n \tau_n + \frac{10 \eta_n^2 \tau_n^4 L^4 (\eta_n \gamma_n)^2}{1 - \mu \tau_n \gamma_n \eta_n + 20 \mu \gamma_n \eta_n^3 L^2 \eta_n^3} \leq 1
    \quad\quad \& \quad\quad
    30 \eta_n^2 L^2 \tau_n^2 \leq 1.
\end{align}
To satisfy the two conditions in \eqref{eq: stepsize cond}, we can pick stepsizes $\eta_n,\gamma_n$ such that
\begin{align} \label{eq: stepsize cond}
    2 \eta_n \gamma_n\tau_n L = 1
    \quad\quad \& \quad\quad
    30 \eta_n^2 L^2 \tau_n^2 \leq 1.
\end{align}
Now we use the result in Lemma \ref{lemma: 3 claims} and put $n = 2m$ to conclude that
\begin{gather} 
    L_n(\w_m) - L_n(\w^*_n) 
    \leq 3 V_{ms},\\
    \Vert \gr L_n (\w_m) \Vert^2 
    \leq 2(2\mu + 1) V_{ms},\\
    \Vert \gr L^{i} (\w_m) \Vert^2 
    \leq 3(2\mu + 1) V_{m s} + 3 V_s. \label{eq: 3 inequalities}
\end{gather}
Substituting the three inequalities \eqref{eq: 3 inequalities} in the sub-optimality error \eqref{eq: fedgate} yields that
\begin{align} \label{eq: fedgate 2}
    \E[L_n(\w) - L_n(\w_n^*)]
    &\leq
    3 \left( 1 - \frac{1}{3} \mu \eta_n \gamma_n \tau_n \right)^{R_n} V_{ms} \\
    &\quad +
    72 \kappa^3 L \tau_n^2 \eta_n^2 \left( (2\mu + 1) V_{m s} + V_s \right)
    +
    48 (2\mu + 1) \kappa L \tau_n^2 \eta_n^2 V_{ms}\\
    &\quad
    +
    24 \kappa^2 L^2 \tau_n^2 \eta_n^3 \sigma^2  +
    15 \kappa L^3 \tau_n^3 \eta_n^2 (\eta_n \gamma_n)^2 \frac{\sigma^2}{n} + \frac{L}{2} \eta_n \gamma_n \frac{\sigma^2}{n}.
\end{align}
We use the fact that $2 \eta_n \gamma_n\tau_n L = 1$ and rearrange the terms in \eqref{eq: fedgate 2} and rewrite it as follows:
\begin{align} \label{eq: fedgate 3}
    \E[L_n(\w) - L_n(\w_n^*)]
    &\leq
    3 \left( 1 - \frac{1}{6 \kappa} \right)^{R_n} V_{ms} \\
    &\quad +
    72 \kappa^3 L \tau_n^2 \eta_n^2 V_s 
    +
    24 (3\kappa^2 + 2) (2\mu + 1) \kappa L \tau_n^2 \eta_n^2 V_{ms}\\
    &\quad
    +
    24 \kappa^2 L^2 \tau_n^2 \eta_n^3 \sigma^2 
    +
    \frac{15}{4} \kappa L \tau_n \eta_n^2 \frac{\sigma^2}{n} 
    +
    \frac{L}{2} \eta_n \gamma_n \frac{\sigma^2}{n}.
\end{align}
To ensure that a model $\w = \w_n$ attains the statistical accuracy of $L_n(\w)$, i.s. $\E[L_n(\w_n) - L_n(\w_n^*)] \leq V_{ns}$, it suffices to have each of the six terms in RHS of \eqref{eq: fedgate 3} less than or equal to $V_{ns}/6$. That is,
\begin{align} \label{eq: 6 inequalities}
    3 \left( 1 - \frac{1}{6 \kappa} \right)^{R_n} V_{ms}
    &\leq 
    \frac{V_{ns}}{6}, \\
    72 \kappa^3 L \tau_n^2 \eta_n^2 V_s 
    &\leq 
    \frac{V_{ns}}{6}, \\
    24 (3\kappa^2 + 2) (2\mu + 1) \kappa L \tau_n^2 \eta_n^2 V_{ms}
    &\leq 
    \frac{V_{ns}}{6}, \\
    24 \kappa^2 L^2 \tau_n^2 \eta_n^3 \sigma^2 
    &\leq 
    \frac{V_{ns}}{6}, \\
    \frac{15}{4} \kappa L \tau_n \eta_n^2 \frac{\sigma^2}{n}
    &\leq 
    \frac{V_{ns}}{6}, \\
    \frac{L}{2} \eta_n \gamma_n \frac{\sigma^2}{n}
    &\leq 
    \frac{V_{ns}}{6},
\end{align}
where $n=2m$ and $V_{ns} = \frac{c}{ns}$ for any $n$. One can check that the following picks for the stepsizes $\eta_n,\gamma_n$ satisfies all the conditions in \eqref{eq: stepsize cond} and \eqref{eq: 6 inequalities}:
\begin{align}
    \eta_n &= \frac{\alpha_n}{\tau_n \sqrt{n}}, \\
    \gamma_n &= \frac{\sqrt{n}}{2 \alpha_n L},
\end{align}
where
\begin{align}    
    \alpha_n 
    \leq
    \min \left\{ 
    \frac{1}{12\sqrt{3} \kappa \sqrt{\kappa L}},
    \frac{\sqrt{n}}{12 \sqrt{2(3\kappa^2 + 2) (2\mu + 1) \kappa L}},
    \left( \frac{ \sqrt{n}}{96 \kappa^2 L^2} \right)^{1/3}, 
    \frac{\sqrt{n}}{\sqrt{15 c \kappa L}} ,
    \frac{\sqrt{n}}{L \sqrt{30}} \right\}.
\end{align}
Moreover, the first and the last conditions in \eqref{eq: 6 inequalities} yield that the number of local updates and the number of communication rounds for the stage with $n$ participating nodes are
\begin{align}    
    \tau_n &= \frac{3}{2} \frac{\sigma^2 s}{c},\\
    R_n &= 12 \kappa \log(6).
\end{align}

\section{Proof of Proposition \ref{prop3}}

In order to characterize the runtime of \texttt{FedGATE}, we first need to determine its two major parameters $\tau$ and $R$. More precisely, we run \texttt{FedGATE} algorithm with all the $N$ available nodes while initialized with arbitrary model $\w_0$ and look for $\tau, R$ after which the global model $\tilde{\w}$ attains the statistical accuracy of $L_N(\w)$, i.e. $ \E[L_N(\tilde{\w}) - L_N(\w_N^*)] \leq V_{Ns}$. We use the convergence guarantee of \texttt{FedGATE} \citep{haddadpour2020federated} in \eqref{eq: fedgate} with $n=N$ nodes, that is,
\begin{align} \label{eq: fedgate N}
    \E[L_N(\w) - L_N(\w_N^*)]
    &\leq
    \left( 1 - \frac{1}{3} \mu \eta \gamma \tau \right)^{R} \left( L_N(\w_0) - L_N(\w_N^*) \right) \\
    &\quad +
    24 \kappa^3 L \tau^2 \eta^2 \frac{1}{N} \sum_{i=1}^{N} \norm{\gr L^i (\w_0)}^2
    +
    24 \kappa L \tau^2 \eta^2 \norm{\gr L_N (\w_0)}^2 \\
    &\quad +
    24 \kappa^2 L^2 \tau^2 \eta^3 \sigma^2  +
    15 \kappa L^3 \tau^3 \eta^2 (\eta \gamma)^2 \frac{\sigma^2}{N} + \frac{L}{2} \eta \gamma \frac{\sigma^2}{N},
\end{align}
where the stepsizes $\eta,\gamma$ satisfy the following conditions:
\begin{align} \label{eq: stepsize cond}
    1 - L \eta \gamma \tau + \frac{10 \eta^2 \tau^4 L^4 (\eta \gamma)^2}{1 - \mu \tau \gamma \eta + 20 \mu \gamma \eta^3 L^2 \eta^3} \leq 1
    \quad\quad \& \quad\quad
    30 \eta^2 L^2 \tau^2 \leq 1.
\end{align}
Note that the initial model $\w_0$ is arbitrary and therefore the initial sub-optimality error can be treated as a constant (and not scaling with $N$), that is, $L_N(\w_0) - L_N(\w_N^*) = \Delta_0$ for a constant $\Delta_0 = \ccalO(1)$. Similarly, we can assume that $\frac{1}{N} \sum_{i=1}^{N} \norm{\gr L^i (\w_0)}^2 = \Delta'_0$ for a constant $\Delta'_0 = \ccalO(1)$ which also yields that $\norm{\gr L_N (\w_0)}^2 \leq \Delta'_0$. We can therefore further simplify \eqref{eq: fedgate N} and write
\begin{align} \label{eq: fedgate N 2}
    \E[L_N(\w) - L_N(\w_N^*)]
    &\leq
    \left( 1 - \frac{1}{3} \mu \eta \gamma \tau \right)^{R} \Delta_0 
    +
    24 \kappa(\kappa^2 + 1) L \tau^2 \eta^2 \Delta'_0 \\
    &\quad +
    24 \kappa^2 L^2 \tau^2 \eta^3 \sigma^2  +
    15 \kappa L^3 \tau^3 \eta^2 (\eta \gamma)^2 \frac{\sigma^2}{N} + \frac{L}{2} \eta \gamma \frac{\sigma^2}{N}.
\end{align}
We furthermore pick the parameters such that $2 \eta \gamma \tau L = 1$ which further simplifies \eqref{eq: fedgate N 2} as follows:
\begin{align} \label{eq: fedgate N 3}
    \E[L_N(\w) - L_N(\w_N^*)]
    &\leq
    \left( 1 - \frac{1}{6 \kappa} \right)^{R} \Delta_0 
    +
    24 \kappa(\kappa^2 + 1) L \tau^2 \eta^2 \Delta'_0 \\
    &\quad +
    24 \kappa^2 L^2 \tau^2 \eta^3 \sigma^2  
    +
    \frac{15}{4} \kappa L \tau \eta^2 \frac{\sigma^2}{N} 
    +
    \frac{L}{2} \eta \gamma \frac{\sigma^2}{N}.
\end{align}
Now to ensure that $ \E[L_N(\tilde{\w}) - L_N(\w_N^*)] \leq V_{Ns}$ holds for a model $\tilde{\w}$ in \eqref{eq: fedgate N 3}, it suffices to satisfy the following inequalities:
\begin{align} \label{eq: inequalities}
    \left( 1 - \frac{1}{6 \kappa} \right)^{R} \Delta_0
    &\leq 
    \frac{V_{Ns}}{5}, \\
    24 \kappa(\kappa^2 + 1) L \tau^2 \eta^2 \Delta'_0
    &\leq 
    \frac{V_{Ns}}{5}, \\
    24 \kappa^2 L^2 \tau^2 \eta^3 \sigma^2
    &\leq 
    \frac{V_{Ns}}{5}, \\
    \frac{15}{4} \kappa L \tau \eta^2 \frac{\sigma^2}{N} 
    &\leq 
    \frac{V_{Ns}}{5}, \\
    \frac{L}{2} \eta \gamma \frac{\sigma^2}{N}
    &\leq 
    \frac{V_{Ns}}{5}, 
\end{align}
with $V_{Ns} = \frac{c}{Ns}$. The following picks for the stepsizes satisfy the aforementioned conditions
\begin{align}
    \eta &= \frac{\alpha}{\tau \sqrt{Ns}}, \\
    \gamma &= \frac{\sqrt{Ns}}{2 \alpha L},
\end{align}
where
\begin{align}    
    \alpha 
    \leq
    \min \left\{ 
    \frac{\sqrt{c}}{2\sqrt{30} \sqrt{\kappa(\kappa^2 + 1) L \Delta'_0}},
    \left( \frac{ \sqrt{Ns}}{96 \kappa^2 L^2} \right)^{1/3}, 
    \frac{\sqrt{Ns}}{\sqrt{15 \kappa L}} ,
    \frac{\sqrt{Ns}}{L \sqrt{30}} \right\}.
\end{align}
Moreover, the number of local updates and the number of communication rounds to reach the final statistical accuracy are as follows:
\begin{align}    
    \tau &= \frac{5}{4} \frac{\sigma^2 s}{c} = \ccalO(s),\\
    R &= 6 \kappa \log \left( \frac{5 \Delta_0 N s}{c} \right) = \ccalO(\kappa \log(N)).
\end{align}
Now note that the expected runtime of each communication round of \texttt{FedGATE} is $\tau T_N$ as the server has to wait for the slowest node that is node $N$ with processing time $T_N$. Therefore, the total expected wall-clock time of \texttt{FedGATE} to reach the final statistical accuracy of all the samples of the $N$ nodes in $L_N(\w)$ is 
\begin{align} 
    \E[T_{\normalfont{\texttt{FedGATE}}}] = R \tau T_N = \ccalO(\kappa s \log(N) T_N), \nonumber
\end{align}
as claimed in Proposition \ref{prop3}.

\section{Proof of Theorem \ref{thm: gain}}

As shown in Proposition \ref{prop2}, the expected runtime for the proposed \texttt{FLANP} with \texttt{FedGATE} in Algorithm \ref{alg: straggler-robust fed} is as follows:
\begin{align} 
    \E[T_{\texttt{FLANP}}] 
    = 
    R_{\texttt{FLANP}} \, \tau_{\texttt{FLANP}}  \sum_{i \,=\, n_0, \, 2n_0, \, 4n_0,\cdots, \, N} T_i
    =
    \frac{18 \log(6)}{c} \kappa s \sigma^2 \left( T_{n_0} + T_{2n_0} + \cdots + T_N \right),
\end{align}
where $R_{\texttt{FLANP}} = 12 \kappa \log(6)$ and $\tau_{\texttt{FLANP}} = 1.5 s \sigma^2 / c$ per Theorem \ref{theorem:1}. Moreover, we showed in Proposition \ref{prop3} that the expected runtime for \texttt{FedGATE} is 
\begin{align} 
    \E[T_{\texttt{FedGATE}}] = R_{\texttt{FedGATE}}  \, \tau_{\texttt{FedGATE}} \, T_N 
    =
    \frac{15}{2c} \kappa s \sigma^2 \log \left( \frac{5 \Delta_0 N s}{c} \right) T_N.
\end{align}
In the case that clients' computation times $T_i$s are random, the expected runtimes are 
\begin{gather} 
    \E[T_{\texttt{FLANP}}] 
    = 
    \frac{18 \log(6)}{c} \kappa s \sigma^2 \left( \E[T_{n_0}] + \E[T_{2n_0}] + \cdots + \E[T_N] \right),\\
    \E[T_{\texttt{FedGATE}}] 
    =
    \frac{15}{2c} \kappa s \sigma^2 \log \left( \frac{5 \Delta_0 N s}{c} \right) \E[T_N]. \label{eq: times}
\end{gather}
Therefore, in order to derive the runtime gain $\frac{\E[T_{\normalfont{\texttt{FLANP}}}]}{\E[T_{\normalfont{\texttt{FedGATE}}}]}$, we first characterize the ratio 
\begin{align} 
    \frac{ \E[T_{n_0}] + \E[T_{2n_0}] + \cdots + \E[T_N] }{\E[T_N]},
\end{align}
where the clients runtimes $T_i$ are i.i.d. with random exponential distribution $\exp(\lambda)$ with rate $\lambda$. Note that we assumed that the clients are sorted with respect to their processing speeds from fastest to slowest. Here, since the computation times $T_i$s are random, we first sort them as $T_{(1)} \leq T_{(2)} \leq T_{(3)} \leq \cdots T_{(N)}$. Without loss of generality and for simplification, let us take $n_0 = 1$ and $\lambda = 1$ and proceed to bound the ratio
\begin{align} \label{eq: ratio}
    \frac{ \E[T_{(1)}] + \E[T_{(2)}] + \E[T_{(4)}] + \cdots + \E[T_{(N)}] }{\E[T_{(N)}]}.
\end{align}
We first provide the following facts about i.i.d. random exponential variables. If $T_i \sim \exp(1)$ are i.i.d. random variables from exponential distribution with mean value of $1$, then the order statistics $T_{(i)}$ have the following properties:
\begin{align} 
    T_{(1)} \sim \exp\left( \frac{1}{N} \right),
    \quad
    T_{(i)} - T_{(i-1)} \sim \exp\left( \frac{1}{N-i+1} \right).
\end{align}
Therefore, the expected value of client $i$'s computation speed $T_{(i)}$ can be written as
\begin{align} 
    \E[T_{(i)}] 
    &=
    \E[T_{(i)} - T_{(i-1)}] + \E[T_{(i-1)} - T_{(i-2)}] + \cdots + \E[T_{(2)} - T_{(1)}] + \E[T_{(1)}] \\
    &=
    \frac{1}{N-i+1} + \frac{1}{N-i+2} + \cdots + \frac{1}{N-1} + \frac{1}{N} \\
    &=
    H_{N} - H_{N-i},
\end{align}
for any $1 \leq n \leq N$. In above, $H_n = 1 + \frac{1}{2} + \frac{1}{3} + \cdots + \frac{1}{n}$ denotes the $n$th harmonic number. Now we use the bounds $\ln(n) + \gamma \leq H_n \leq \ln(n+1) + \gamma$ for each $n\geq 2$ where $\gamma \approx 0.577$ is the Euler-Mascheroni constant. For further simplification, we assume that $N$ is a power of $2$, that is $N=2^K$ for some integer $K$. Therefore, we can write
\begin{align} 
    \E[T_{(1)}] 
    &=
    H_{N} - H_{N-1}
    \leq
    \ln(N+1) - \ln(N-1),\\
    \E[T_{(2)}] 
    &=
    H_{N} - H_{N-2}
    \leq
    \ln(N+1) - \ln(N-2),\\
    \E[T_{(4)}] 
    &=
    H_{N} - H_{N-4}
    \leq
    \ln(N+1) - \ln(N-4),\\
    \E[T_{(8)}] 
    &=
    H_{N} - H_{N-8}
    \leq
    \ln(N+1) - \ln(N-8),\\
    \vdots\\
    \E[T_{(N/2)}] 
    &=
    H_{N} - H_{N/2}
    \leq
    \ln(N+1) - \ln(N/2),\\
    \E[T_{(N)}] 
    &=
    H_{N}
    \leq
    \ln(N+1) + \gamma.
\end{align}
Therefore, we can bound the numerator of the ratio in \eqref{eq: ratio} as follows:
\begin{align} \label{eq: numerator}
    &\E[T_{(1)}] + \E[T_{(2)}] + \E[T_{(4)}] + \cdots + \E[T_{(N)}] \\
    &\quad\leq
    (K+1) \ln\left(2^K + 1\right) + \gamma - \ln\left( \left(2^K-1\right) \left(2^K-2\right) \left(2^K-4\right)\cdots \left(2^{K-1}\right) \right) \\
    &\quad \leq
    (K+1) \ln\left(2^K + 1\right) + \gamma - (K^2 - K) \ln(2) \\
    &\quad \leq
    (K+1) \left(K \ln(2) + \frac{1}{2^K}\right) + \gamma - (K^2 - K) \ln(2) \\
    &\quad =
    K \left(2 \ln(2) + \frac{1}{2^K}\right) + \frac{1}{2^K} + \gamma
\end{align}
Moreover, the denominator of the ratio in \eqref{eq: ratio} can be bounded as follows:
\begin{align} \label{eq: denom}
    \E[T_{(N)}] 
    =
    H_{N}
    \geq \ln(N) + \gamma
    =
    K \ln(2) + \gamma.
\end{align}
Putting \eqref{eq: numerator} and \eqref{eq: denom} together, we can bound the ratio in \eqref{eq: ratio} as follows:
\begin{align} \label{eq: ratio 2}
    \frac{ \E[T_{(1)}] + \E[T_{(2)}] + \E[T_{(4)}] + \cdots + \E[T_{(N)}] }{\E[T_{(N)}]}
    \leq
    \frac{K \left(2 \ln(2) + \frac{1}{2^K}\right) + \frac{1}{2^K} + \gamma}{K \ln(2) + \gamma}
    \leq 2 + \frac{1}{N}.
\end{align}
Now, we are able to precisely characterize the speedup gain of \texttt{FLANP} compared to \texttt{FedGATE} according to the expressions in \eqref{eq: times} and the ratio in \eqref{eq: ratio 2} to conclude that
\begin{align} 
    \frac{\E[T_{\texttt{FLANP}}]}{\E[T_{\texttt{FedGATE}}] }
    &= 
    \frac{12 \log(6)}{5  \log \left( {5 c^{-1} \Delta_0 N s} \right)} \frac{ \E[T_{(1)}] + \E[T_{(2)}] + \E[T_{(4)}] + \cdots + \E[T_{(N)}] }{\E[T_{(N)}]} \\
    &\leq
    \frac{12 \log(6)}{5  \log \left( {5 c^{-1} \Delta_0 N s} \right)} \left(2 + \frac{1}{N} \right)\\
    &=
    \ccalO\left( \frac{1}{\log(Ns)} \right),
\end{align}
which completes the proof of Theorem \ref{thm: gain}.

\end{document}